\pdfoutput=1

\documentclass[11pt]{article}

\usepackage{multirow}

\usepackage{acl}

\usepackage{times}
\usepackage{latexsym}

\usepackage{comment}

\usepackage{algorithm}
\usepackage{algorithmic}
\usepackage{placeins}

\usepackage{multirow}
\usepackage{tikz}
\usetikzlibrary{shapes.geometric, arrows, positioning}
\usepackage{xcolor}
\usepackage{fancyvrb}
\usepackage{pgfplots}
\usepackage{array}
\usepackage{tabularx}
\usepackage{float}
\usepackage{pgf-pie}
\usepackage{subcaption}
\usepackage{enumitem}
\usepackage{setspace}
\usepackage{booktabs} 
\usepackage{siunitx} 
\usepackage{hyperref}

\usepackage{pifont}
\newcommand{\cmark}{\ding{51}}%
\newcommand{\xmark}{\ding{55}}%
\newcommand{\up}{\ding{115}}%
\newcommand{\down}{\ding{116}}%

\newcommand{\uptick}{{\color{darkgreen}$\uparrow$}}
\newcommand{\downtick}{{\color{darkred}$\downarrow$}}
\definecolor{darkgreen}{rgb}{0.0, 0.5, 0.0}
\definecolor{darkred}{rgb}{0.55, 0.0, 0.0}

\usepackage[T1]{fontenc}

\usepackage[utf8]{inputenc}

\usepackage{microtype}

\usepackage{inconsolata}

\usepackage{adjustbox}

\title{Confidence Under the Hood: An Investigation into the Confidence-Probability Alignment in Large Language Models}

\author{%
  Abhishek Kumar\textnormal{\textsuperscript{1},}
  Robert Morabito\textnormal{\textsuperscript{1},} 
  Sanzhar Umbet\textnormal{\textsuperscript{2},}
  Jad Kabbara\textnormal{\textsuperscript{3},}
  \textnormal{and}
  Ali Emami\textsuperscript{\textnormal{1}} \\
  \textsuperscript{1}Brock University, St. Catharines, ON, Canada \\
  \textsuperscript{2}Nazarbayev University, Astana, Kazakhstan \\
  \textsuperscript{3}Massachusetts Institute of Technology, Cambridge, MA, USA \\
\texttt{\{aa22dt, rm20mg, aemami\}@brocku.ca} \\
\texttt{sanzhar.umbet@alumni.nu.edu.kz}\\
  \texttt{jkabbara@mit.edu}
}

\begin{document}
\maketitle
\begin{abstract}
As the use of Large Language Models (LLMs) becomes more widespread, understanding their self-evaluation of confidence in generated responses becomes increasingly important as it is integral to the reliability of the output of these models. We introduce the concept of Confidence-Probability Alignment, that connects an LLM's internal confidence, quantified by token probabilities, to the confidence conveyed in the model's response when explicitly asked about its certainty. Using various datasets and prompting techniques that encourage model introspection, we probe the alignment between models' internal and expressed confidence. These techniques encompass using structured evaluation scales to rate confidence, including answer options when prompting, and eliciting the model's confidence level for outputs it does not recognize as its own. Notably, among the models analyzed, OpenAI's GPT-4 showed the strongest confidence-probability alignment, with an average Spearman's $\hat{\rho}$ of 0.42, across a wide range of tasks. Our work contributes to the ongoing efforts to facilitate risk assessment in the application of LLMs and to further our understanding of model trustworthiness.\footnote{The code to reproduce our experimental results as well as detailed interactions with the language models are available at \href{https://github.com/akkeshav/confidence\_probablitiy\_alignment}{https://github.com/akkeshav/confidence\_probablitiy\_alignment}.}
\end{abstract}

\section{Introduction}
\label{sec:Intro}

In recent years, we have witnessed the rapid development and deployment of Large Language Models (LLMs) across various disciplines. LLMs such as GPT \citep{NEURIPS2020_1457c0d6, schulman2022chatgpt, openai2023gpt4}, PaLM \citep{chowdhery2022palm}, Chincilla \cite{hoffmann2022training}, and LLaMa \citep{touvron2023llama}, have showcased remarkable performance across a diverse range of NLP tasks, and their capabilities in empowering chatbots have ignited a surge of interest among the general public. With the ongoing integration of these models into high-stakes areas such as healthcare, law, and education, a critical evaluation of their behavior and trustworthiness is becoming increasingly essential.

\begin{figure}[t]
    \centering
    \includegraphics[width =0.8\linewidth]{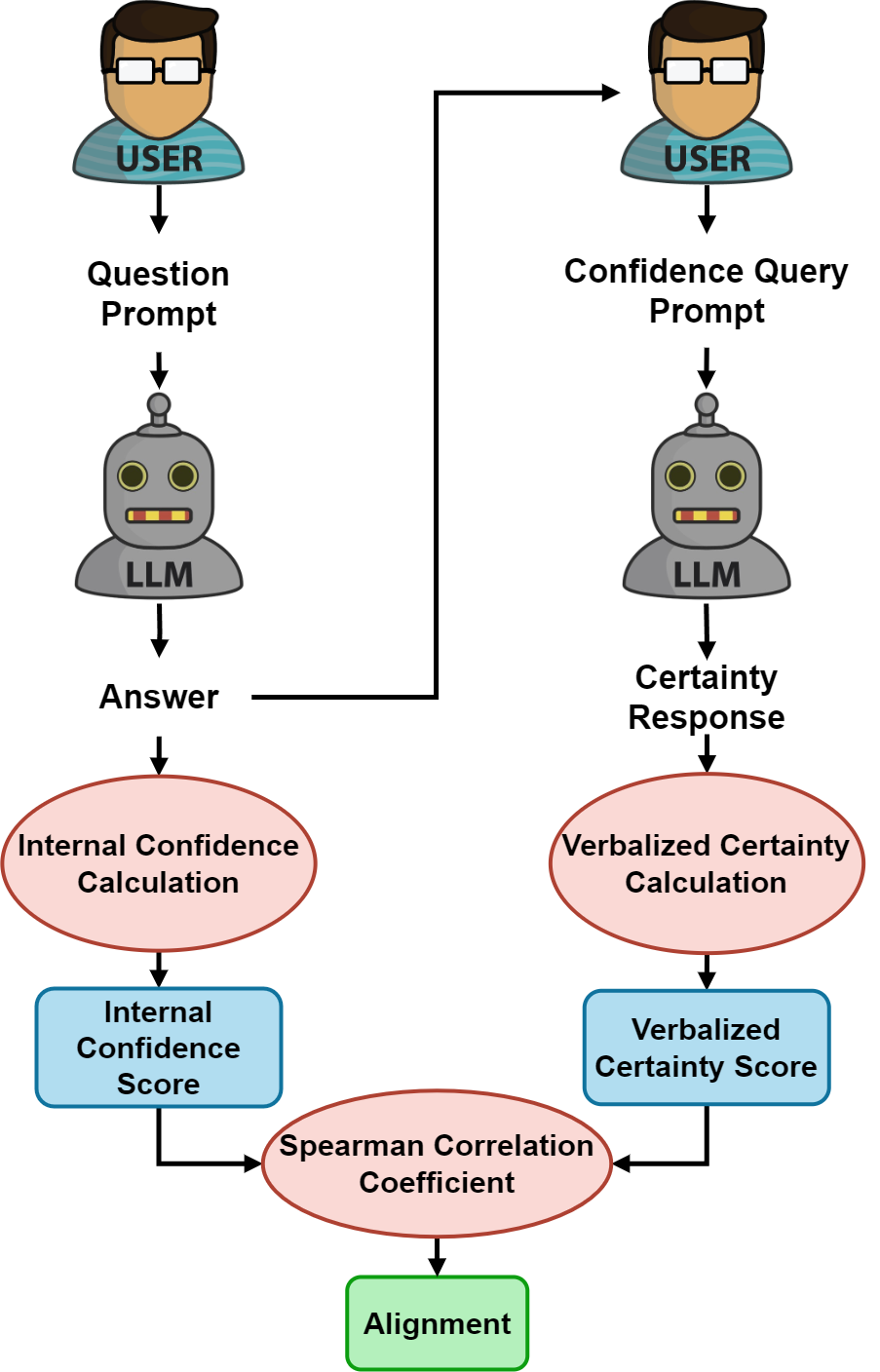}
    \caption{Flow diagram illustrating the process of extracting and comparing the Internal Confidence and Verbalized Certainty in an LLM.}
    \label{fig:flowdiagram}
\end{figure}

\begin{figure*}[htp]
    \centering
    \includegraphics[width=0.9\linewidth]{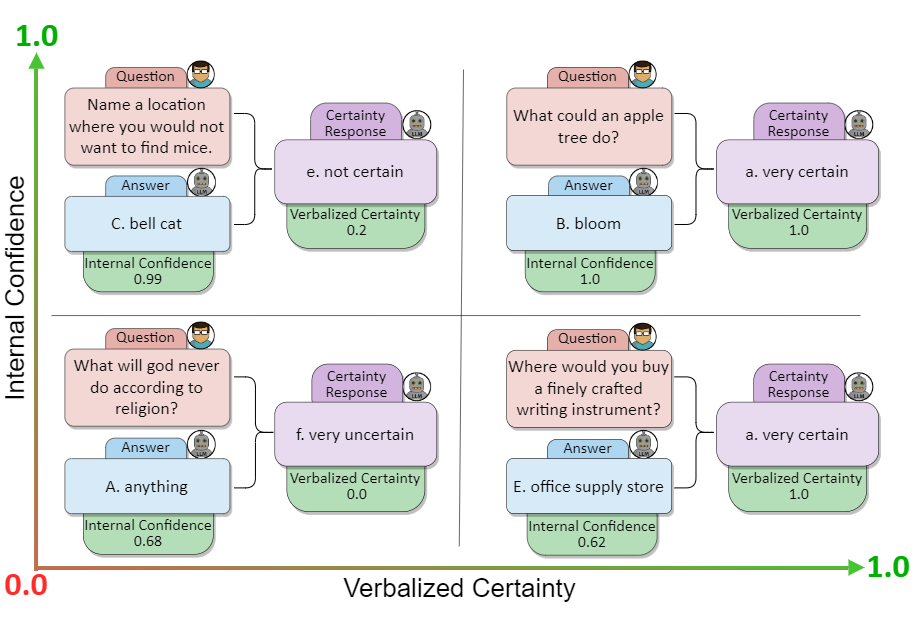}
    \caption{Illustration of \textbf{GPT-4}'s responses to various questions, accompanied by their internal confidences and expressed certainty levels. Questions sourced from CommonsenseQA dataset.}
    \label{fig:intro-chart}
\end{figure*}

Many contemporary prompting techniques, such as Self-Consistency \citep{wang2023selfconsistency}, Tree of Thoughts \citep{yao2023tree}, and Multi-Agent Debating \citep{du2023improving}, rely heavily on a model's self-evaluation of its reasoning process. Recent works like \cite{diao2023active} use LLMs' self-confidence for tasks like question selection based on uncertainty. However, if a misalignment exists between the model's expressed self-reasoning and its true internal confidence, these techniques could yield misleading results, undermining their practical utility. Trust and decision-making by users of these models in real-world applications are directly impacted by understanding this alignment.

This is further complicated by LLM hallucination \cite{huanghalluc, shuster2021retrieval,hallucsurvey}, where LLMs produce outputs that seem plausible but are factually incorrect or fabricated. These hallucinations often come with high expressions of confidence, challenging non-expert users to distinguish them from reliable outputs. Notable examples of these are reported in \cite{alkaissi2023artificial}, where ChatGPT confidently cited non-existent references for specific medical claims.

Amid these challenges, our work aims to enhance trustworthiness in LLMs through a detailed understanding of model confidence. We define verbalized certainty as an LLM's explicit expression of its confidence level in response to a question and investigate its relationship with the model's actual internal confidence, quantified by token probabilities. Our exploration of the correlation between these two measures, termed \textit{Confidence-Probability Alignment}, posits that this alignment is crucial for the reliability of a model's output. 

Drawing upon a range of question datasets, our investigation explores the nuances of internal and verbalized confidence across an extensive array of models, including OpenAI's GPT-3, its variants such as InstructGPT and the RLHF version, and the more recent GPT-4, in addition to open-source models, namely Microsoft's Phi-2-2.7B and HuggingFace's Zephyr-7B. We probe the models with questions from a variety of tasks, gather response token probabilities, and solicit confidence in the answers  (visually summarized in Figure \ref{fig:flowdiagram}). We also explore model response variability under different parameters and prompts and examine the connection between a model's confidence and its actual performance.  Figure \ref{fig:intro-chart} provides a visual overview of how GPT-4 responds to different questions and its associated expressions of confidence. Our contributions are three-fold:

\paragraph{1. Conceptualization of Confidence-Probability Alignment:} We introduce a novel framework to assess LLMs' transparency and reliability. This measure evaluates the correlation between an LLM's internal confidence and its verbalized certainty, offering a new lens to better understand model behavior. 

\looseness=-1\paragraph{2. Study Across Diverse LLM Architectures:} Our work covers a wide array of LLM architectures. Using prompting techniques to encourage model introspection, we reveal varied alignment dynamics, notably GPT-4's consistent confidence-probability alignment across multiple tasks.

 \paragraph{3. Analysis of Confidence-Correctness Relationship:} We further analyze the link between models' expressed confidence and response accuracy, exploring the impact of model-specific parameters such as temperature. Additionally, we develop a taxonomy of misalignments for detailed error analysis, uncovering various cases of alignment discrepancies and their implications for model reliability.

\section{Confidence-Probability Alignment}
\label{sec:ConfProbAlign}
Confidence-Probability Alignment refers to the correlation between a model's verbalized certainty and its internal confidence, each quantified in specific ways. We define verbalized certainty as a model's explicit expression (i.e., in the generated text) of its confidence level regarding its own answer to a given question. The internal confidence, on the other hand, is quantified using answer token probabilities. The alignment of these two aspects can be determined using the following procedures.

\subsection{Response Generation}
The generation of a response from the language model commences with a structured prompt, comprising a question denoted as $Q$, and a set of answer options, hereafter denoted as $O_{set}$. Each individual answer option within $O_{set}$ will be referred to as $O_i$. The answer options in $O_{set}$ are potential responses that the model can select to answer $Q$, with this structure designed to elicit a definitive response.

For instance, consider a question $Q$ posed as ``Which of the following is a common element in the atmosphere?"  The corresponding set of answer options $O_{set}$ comprises five distinct responses labeled from `A' to `E': A. Oxygen, B. Nitrogen, C. Gold, D. Iron, and E. Helium.

The question and its options are concatenated in a structured format, each answer option prefixed with a corresponding label, and separated by newline characters to maintain a clear distinction. The prompt culminates with the term `Answer:', designed to solicit a model's response.

The prompt is constructed as follows:

\begin{verbatim}
Prompt = Q + "\n" + Option A + "\n" + ...
+ Option E + "\nAnswer: "
\end{verbatim}

Upon presenting this structured prompt to a language model, the model generates a response text. We subsequently extract the chosen answer $a_i$ from the output response text. This chosen answer, $a_i \in O_{set}$, symbolizes the model's selected response from the options $O_{set}$, serving as the basis for the ensuing evaluation of the model's internal confidence and verbalized certainty.

\subsection{Internal Confidence}

In the context of LLMs, internal confidence for a chosen output is quantified as the probability assigned to the selected output token, $T_{\mathrm{choice}}$. This probability can be derived by applying the exponential function to the log probabilities or logits of each output token, depending on the model's output format. A higher score for $T_{\mathrm{choice}}$ signifies greater model confidence in that choice. However, exact confidence assessment can be challenging due to potential token ambiguities, such as differing case sensitivity ('B' vs. 'b') for answer tokens.\footnote{Although accounted for, we observed this phenomenon to occur in less than three percent of instances across datasets.}

To address this, we introduce the concept of \textit{adjusted} answer token probability. This involves calculating token probabilities and adjusting them to account for potential ambiguities and variations in token format, which we detail in Algorithm \ref{alg:adjusted_token_prob}.

Our algorithm starts by converting log probabilities (or, alternatively, logits) to standard probabilities for all tokens. Then, for each answer option, it identifies all corresponding tokens, calculating the highest probability among them to represent the option's most confident token representation. Finally, the algorithm normalizes the highest probability of the selected answer option against the sum of probabilities for all options, yielding an adjusted internal confidence measure, $P_{IC}$.

\textbf{GPT-Family Models:} For GPT-family models, which return log probabilities ($\log P(T_i)$), token probabilities are derived using:

\begin{equation}
P(T_i) = \exp(\log P(T_i)).
\end{equation}

\textbf{Open-Source Models:} Many open-source models provide logits ($L(T_i)$) for output tokens. To convert these logits into probabilities, we employ the softmax function:

\begin{equation}
P(T_i) = \frac{e^{L(T_i)}}{\sum_{j} e^{L(T_j)}}.
\end{equation}

\begin{algorithm}[t]
\small
\caption{Procedure for computing adjusted token probabilities, $P_A$ and internal confidence, $P_{IC}$}
\label{alg:adjusted_token_prob}
\begin{algorithmic}[1]

\REQUIRE Model output tokens $T$ and their corresponding log probabilities $\log P(T)$.
\STATE Compute token probabilities, $P(T_i)$, for all tokens $T_i$ in $T$ using $P(T_i) = \exp(\log P(T_i))$.
\FOR{each Option $O$ in the list of possible options}
    \STATE Determine the set of tokens, $T_O$, corresponding to option $O$.
    \STATE Compute the maximum probability among tokens in $T_O$ to obtain $P_{O} = \max\limits_{i \in T_O} P(T_i)$.
    
    \STATE Append $P_{O}$ to the set of adjusted probabilities $P_A$.
\ENDFOR
\STATE Select maximum probability, $P_M$ from $P_A$
\STATE $P_S \leftarrow 0$.
\FOR{each probability $P_I$ in $P_A$}
    \STATE $P_S \leftarrow P_S + P_I$ \\
\ENDFOR
\STATE Compute internal confidence $P_{IC}$, by dividing $P_M$ by $P_S$.
\RETURN $P_{IC}$
\end{algorithmic}
\end{algorithm}

\subsection{Verbalized Certainty}

We define verbalized certainty as the language model's explicit expression of its confidence level in its own answer to a given question. To extract this measure, we construct a Confidence Querying Prompt (CQP) that asks the model to analyze its certainty in the context of the question, the available options, and its chosen answer. The CQP is presented as follows:

\definecolor{darkred}{rgb}{0.8, 0.0, 0.0}

\definecolor{darkgreen}{rgb}{0.0, 0.5, 0.0}

\definecolor{darkblue}{rgb}{0.0, 0.0, 0.8}

\DefineVerbatimEnvironment{ColorVerbatim}{Verbatim}{formatcom=\color{darkred}}

\begin{ColorVerbatim}
A language model was asked: {question}
\end{ColorVerbatim}

\vspace{-5mm}
\DefineVerbatimEnvironment{ColorVerbatim}{Verbatim}{formatcom=\color{darkgreen}}
\begin{ColorVerbatim}
Options were: {choices_text}.
\end{ColorVerbatim}
\vspace{-4mm}
\begin{verbatim}
The model's answer was: {response_text}.

Analyse its answer given other options. 
How certain are you of the model's answer?
\end{verbatim}
\vspace{-4mm}
\DefineVerbatimEnvironment{ColorVerbatim}{Verbatim}{formatcom=\color{darkblue}}
\begin{ColorVerbatim}
a. Very Certain
b. Fairly Certain
c. Moderately Certain
d. Somewhat Certain
e. Not Certain
f. Very Uncertain
\end{ColorVerbatim}

This query effectively asks the model to take on an observational role and to analyze the certainty of its answer. The model's response gives us its verbalized certainty. Below, we outline the key components of the CQP. 

\paragraph{\textbf{Simulation of Third-Person Perspective \textcolor{darkred}{(TPP)}:}}
The CQP initiates with a third-person perspective (TPP) to mitigate potential self-preferential biases \cite{wang2023large,zheng2023judging}. It presents the question and answer as if they may not have necessarily been generated by itself. In this scenario, the model's `confidence' reflects its probabilistic estimation of the response's correctness, based on its training. It's important to clarify that this does not equate to personal confidence or self-awareness, but rather an approximation of answer accuracy.
Ultimately, the TPP helps us elicit a measure of confidence that is less susceptible to potential biases, which have been noted in the mentioned studies, arising from the language model's own generation process.

\paragraph{\textbf{Option Contextualization \textcolor{darkgreen}{(OC)}:}}
The Option Contextualization (OC) aspect of the CQP equips the model with a framework for gauging its chosen response by stating, `Options were: \{choices\_text\}'. This  accomplishes two goals:

\textit{Comparative Evaluation:} By displaying all options, the model can contextualize its chosen response, allowing for more informed confidence judgments compared to isolated evaluations.

\textit{Answer Verification:} Providing all potential answers facilitates comprehensive evaluations and enables the model to adjust its confidence if the selected answer is sub-optimal in comparison to the other options.

\paragraph{\textbf{Likert Scale Utilization \textcolor{darkblue}{(LSU)}:}}
The Likert Scale Utilization (LSU) phase of the CQP employs a qualitative scale ranging from `very certain' to `very uncertain'. The choice of a qualitative scale instead of a numerical one aims to maintain a consistent understanding across different model instances. In the context of LLMs, the interpretation of a numerical certainty scale (e.g., `rate your certainty from 1 to 10') can vary significantly between different model instances due to the lack of concrete experiential or emotional grounding that humans use to interpret these numbers. A qualitative scale leverages the model's training on a vast corpus of human language, which makes the gradations (e.g., `fairly certain' is more confident than `moderately certain') generally more universally understood and consistent.\\
\indent Responses to the CQP are then mapped to a numerical score using a predefined system: `very certain' equals 1.0, `fairly certain' equals 0.8, `moderately certain' equals 0.6, `somewhat certain' equals 0.4, and `not certain' or `very uncertain' correspond to 0.2 and 0, respectively. This scoring method provides a quantifiable measure of the model's verbalized certainty. Importantly, this measure reflects the model's self-evaluated certainty and is distinct from its internal confidence as quantified by the adjusted token probabilities. \\
\indent The finalized CQP design emerged after exploring various alternative designs during preliminary experiments, as detailed in Appendix Table \ref{tab:Prompt Approaches}. 

\begin{table*}[h]
  \centering
  \renewcommand{\arraystretch}{1.2}
  \footnotesize
  \setlength{\tabcolsep}{5pt}
  \begin{tabular}{lccccc} 
    \toprule
    Model & CSQA & QASC & Riddle Sense & OpenbookQA & ARC \\
    \midrule
    GPT-4 (gpt-4-0613) & \textbf{0.42} & \underline{\textbf{0.47}} & \textbf{0.42} & \textbf{0.41} & \textbf{0.35} \\
    InstructGPT-3 + RLHF (text-davinci-003) & 0.40 & 0.40 & 0.35 & 0.25 & 0.25 \\
    InstructGPT-3 (text-davinci-002) & 0.15 & 0.16 & 0.19 & 0.13 & 0.17 \\
    GPT-3 (text-davinci-001) & 0.01 & -0.01 & -0.01 & -0.02 & -0.04 \\
    Zephyr-7B & -0.14 & -0.10 & 0.02 & 0.12 & 0.15 \\
    Microsoft-Phi-2 & -0.02 & -0.05 & 0.00 & 0.03 & 0.07 \\
    \bottomrule
  \end{tabular}
  \caption{Alignment evaluation using \textbf{Spearman's rank correlation coefficient}. All values are significant ($p<0.01$). Highest value for each dataset in \textbf{bold}.}
\label{tab:Spearman's rank}
\end{table*}

\subsection{Alignment Evaluation}
We define alignment as the correlation between internal and verbalized confidence metrics, evaluated using Spearman's rank correlation coefficient.

\paragraph{\textbf{Spearman's Rank Correlation Coefficient:}}
The Spearman's rank correlation coefficient is a non-parametric test that measures the degree of association between two variables. Unlike the Pearson correlation coefficient, which requires the assumption of normally distributed variables, Spearman's correlation does not require this assumption and can handle ordinal, interval, and ratio data. This makes it ideal for comparing the non-normally distributed token probabilities and verbalized confidence values in our study.

Given two variables $X$ and $Y$, the Spearman correlation $\rho$ is computed as follows:

\begin{equation}
\vspace{2mm}
\rho = 1 - \frac{6 \sum d_i^2}{n(n^2 - 1)},
\end{equation}

where $d_i$ is the difference between the ranks of corresponding values of $X$ and $Y$, and $n$ is the number of observations.

\section{Experiments}
\subsection{Tasks \& Datasets}
We select a diverse set of tasks that allow us to assess an LLM’s confidence-probability alignment across various question types and complexities.

\begin{itemize}[itemsep=1pt, parsep=1pt]
\item \textbf{CommonsenseQA (CSQA)} \cite{talmor2019commonsenseqa}: CSQA poses multiple-choice questions that require a grasp of common-sense knowledge to answer effectively.
\item \textbf{Question Answering via Sentence Composition (QASC)} \cite{Khot2019QASCAD}: QASC necessitates models to infer answers by composing information sourced from multiple sentences, thereby testing their ability to form connections.
\item \textbf{RiddleSense} \cite{lin2021riddlesense}: The unique dataset of RiddleSense tests models on riddles, challenging them to use a blend of world knowledge and lateral thinking.
\item \textbf{OpenBookQA} \cite{Mihaylov2018CanAS}: OpenBookQA tests models on science-based multiple choice questions, gauging their understanding of factual scientific information.
\item \textbf{AI2 Reasoning Challenge (ARC)} \cite{clark2018think}: The ARC challenges models with complex, knowledge-intensive questions that necessitate deep reasoning and the integration of multiple information sources, far beyond simple question answering.
\end{itemize}

\subsection{Models}
We used the following LLMs: \textbf{GPT-3 (\textit{text-davinci-001})} \cite{NEURIPS2020_1457c0d6}; \textbf{InstructGPT-3 (\textit{text-davinci-002})} \cite{ouyang2022training}; \textbf{InstructGPT-3 + RLHF (\textit{text-davinci-003})} \cite{ouyang2022training}; \textbf{GPT-4 (\textit{gpt-4-0613}; \cite{openai2023gpt4})}; \textbf{Microsoft's Phi-2-2.7B}\footnote{https://ai.azure.com/explore/models/microsoft-phi-2/version/4/registry/azureml-msr} and \textbf{HuggingFace's Zephyr-7B} \cite{tunstall2023zephyr}.\footnote{Selected for their balance between computational efficiency and answer quality, these open-source models offer an ideal compromise—being sufficiently compact for algorithmic tractability while robust enough to deliver meaningful responses for question answering and verbalized certainty evaluations.}

\subsection{Experimental Design} We prompted models with the complete set of all dataset questions, obtained the responses, and extracted answer token  probabilities for internal confidence estimation (following Algorithm \ref{alg:adjusted_token_prob}). We then prompt the models (using CQP) to verbally express their confidence to compute alignment with their internal confidence.  For a complete visual walkthrough of the procedure through an example, please refer to Appendix Figure \ref{fig:fullProcess}.\footnote{To conduct all experiments, approximately 132.5 compute hours were elapsed. For experiments with GPT models, their public OpenAI API was used; https://openai.com/api/.}

\definecolor{myblue}{RGB}{66,133,244}
\definecolor{mygreen}{RGB}{76,175,80}
\definecolor{myorange}{RGB}{255,152,0}
\definecolor{mypurple}{RGB}{156,39,176}
\definecolor{myred}{RGB}{244,67,54}
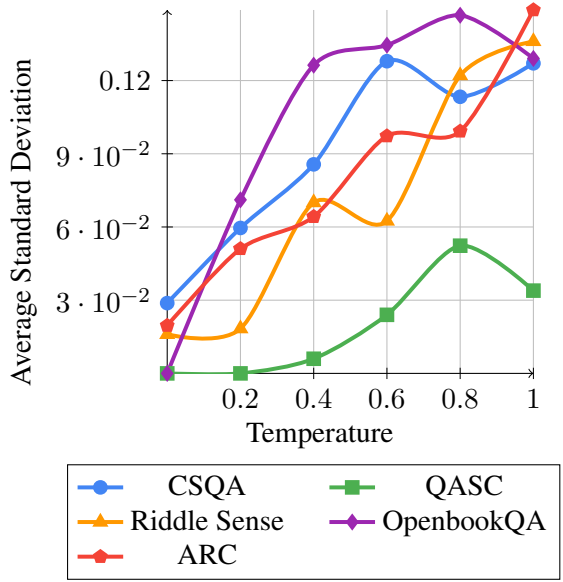
\begin{figure}[htp!]

\centering
\begin{tikzpicture}

\begin{axis}[
    grid=major,
    width=0.40\textwidth,
    height=0.40\textwidth,
    axis lines=middle,
    axis line style={->},
    tick style={color=black},
    xtick distance=0.2,
    ytick distance=0.03,
    every axis plot/.append style={line width=1.5pt},
    legend style={
        at={(0.4, -0.25)},   
        anchor=north, 
        legend columns=2,
        /tikz/every even column/.append style={column sep=0.5cm}
    },
]
\addplot[color=myblue,mark=*,smooth] coordinates {
    (0, 0.02881)
    (0.2, 0.0596)
    (0.4, 0.08566)
    (0.6, 0.12793)
    (0.8, 0.11335)
    (1, 0.12706)
};
\addlegendentry{CSQA}

\addplot[color=mygreen,mark=square*,smooth] coordinates {
    (0, 0)
    (0.2, 0)
    (0.4, 0.006)
    (0.6, 0.024)
    (0.8, 0.05233)
    (1, 0.03392)
};
\addlegendentry{QASC}

\addplot[color=myorange,mark=triangle*,smooth] coordinates {
    (0, 0.016)
    (0.2, 0.01833)
    (0.4, 0.07)
    (0.6, 0.06245)
    (0.8, 0.12205)
    (1, 0.13604)
};
\addlegendentry{Riddle Sense}

\addplot[color=mypurple,mark=diamond*,smooth] coordinates {
    (0, 0)
    (0.2, 0.07112)
    (0.4, 0.1263)
    (0.6, 0.13452)
    (0.8, 0.14676)
    (1, 0.12906)
};
\addlegendentry{OpenbookQA}

\addplot[color=myred,mark=pentagon*,smooth] coordinates {
    (0, 0.0196)
    (0.2, 0.05107)
    (0.4, 0.06421)
    (0.6, 0.09727)
    (0.8, 0.09922)
    (1, 0.14892)
};
\addlegendentry{ARC}

\end{axis}
\node[below, yshift=-15pt, xshift = -80] at (current axis.right of origin) {Temperature};
\node[rotate=90, above,  yshift=45pt, xshift = -80] at (current axis.above origin) {Average Standard Deviation};
\end{tikzpicture}

\caption{Comparative 
analysis illustrating the relationship between temperature and average standard deviation of verbalized certainty across different datasets.}

\label{fig:Temperature Stability}
\end{figure}
\stepcounter{figure}

\section{Results}
\label{sec:results}

\paragraph{Alignment Evaluation}
Table \ref{tab:Spearman's rank} presents the dataset-wise alignment evaluation using Spearman's rank correlation coefficient. Here, GPT-4 consistently outperforms its counterparts across every dataset. Specifically, it registered the highest coefficient on the QASC datasets of nearly 0.5, indicative of a moderate correlation. In contrast, OpenbookQA and ARC datasets marked the lowest correlations, although they still remained higher than other model versions with values of 0.41 and 0.35, respectively.

The standout performance of GPT-4 highlights potential advancements in model architecture and training methodologies, including the scale and possibly more refined human feedback integration.

\newpage
\begin{table*}[htp!]
\small
  \centering
  \newcolumntype{C}{>{\centering\arraybackslash}X}
  \begin{tabularx}{\textwidth}{p{1.6cm} p{1.8cm} | X | c}
    \hline
    \textbf{Int. Conf.} & \textbf{Verb. Cert.} & \textbf{Example} & \textbf{Alignment Type} \\ \hline
     & & \textbf{Q}: Where would you expect to find white mice? & Consistent Alignment \\
     \textcolor{green}{\up} 1.00 & \textcolor{green}{\up} 1.00 & \textbf{A}: E. laboratory & \\
    \hline
     & & \textbf{Q}: Where did you meet your best friend since Kindergarten? & Internal Overconfidence \\
     \textcolor{green}{\up} 0.99 & \textcolor{red}{\down} 0.20 & \textbf{A}: B. School & \\
    \hline 
     & & \textbf{Q}: Where would you be concerned about finding a cavity? & External Overconfidence \\
     \textcolor{red}{\down} 0.69 & \textcolor{green}{\up} 1.00 & \textbf{A}: E. teeth & \\
    \hline 
     & & \textbf{Q}: What is likely to satisfy someone's curiosity? & Consistent Discordance \\
     \textcolor{red}{\down} 0.69 & \textcolor{red}{\down} 0.40 & \textbf{A}: A. hear news & \\
    \hline
  \end{tabularx}
  \caption{Instances of similar and contrasting values of internal confidence (Int. Conf.) and verbalized certainty (Verbal Cert.). \textbf{Q} represents the model's question, and \textbf{A} is the model's answer. Tests are from \textbf{GPT-4} (\textit{gpt-4-0613}) using the CommonsenseQA dataset. For examples using \textbf{InstructGPT-3 + RLHF}, refer to Appendix Table \ref{tab:Illustrative Examples GPT3}.}
  \label{tab:Illustrative Examples}
\end{table*}

\setlength{\tabcolsep}{5pt}

\begin{table*}[h]
\small
  \centering
  \newcolumntype{C}{>{\centering\arraybackslash}X}
  \begin{tabularx}{\textwidth}{c | c | c | c}
    \hline
    \textbf{Question}  & \textbf{Verb. Certainty Response} & \textbf{Model} & \textbf{Comment}\\ \hline
     \textbf{Q}: What happens to light rays in water? & g. Fairly certain, h. Moderately certain & Phi & Unable to choose one \\
     \textbf{A}: B. Refraction & i. Somewhat certain, j. Not certain & & correct certainty response \\ \hline
     
    \textbf{Q}: An example of an animal that has  & DNA sequence. because most carcinogens & Phi & Response lacks verbalized \\
     permeable skin is? \textbf{A}: G. Frog &  cause cancer by altering the DNA .. & & certainty option \\ \hline

     \textbf{Q}: The work of what causes a river to   & Model provided a clear explanation & Zephyr & Response lacks verbalized \\
     become deeper ? \textbf{A}: F. Erosion &  for why F is the correct .. & & certainty option \\ \hline

      \textbf{Q}: Cell phones can cause people in cars & Other options were: A. eardrum & Zephyr & The response merely   \\
     to what? \textbf{A}: B. Distraction &  C. death, D. Injury, E. stop for gas .. & & reiterates the initial options. \\ \hline
     
    \hline
  \end{tabularx}
  \caption{Instances demonstrating subpar performance of small open-source models (Microsoft's Phi and Zephyr \cite{tunstall2023zephyr}). Each row highlights a failure case of these models to generate verbalized certainty from the QASC dataset. \textbf{Q} represents the model’s question, and \textbf{A} is the model’s answer, and are provided as part of the Confidence Querying Prompt.}
  \label{tab:Failure illustrative Examples}
\end{table*}

\begin{figure*}[h]

    \centering
    \begin{tikzpicture}[scale=0.35]
 
        \begin{scope}[shift={(0,4)}]
            \node at (2.5, 6) {\textbf{CSQA}};
            \node at (1.25,4.5) {\cmark};
            \node at (3.75,4.5) {\xmark};
            \node at (0,3.25) {\textbf{+}};
            \node at (0,1.75) {\textbf{-}};
            
            \draw (0.5,1) rectangle (4.5,4); 
            \draw (0.5,2.5) -- (4.5,2.5); 
            \draw (2.5,1) -- (2.5,4); 
            
            \node at (1.5,3.25) {\textbf{\fontsize{9.5}{14}\selectfont 64}};
            \node at (3.5,3.25) {\textbf{\fontsize{9.5}{14}\selectfont 8.4}};
            \node at (1.5,1.75) {\textbf{\fontsize{9.5}{14}\selectfont 16.8}};
            \node at (3.5,1.75) {\textbf{\fontsize{9.5}{14}\selectfont 8.8}};

            \filldraw[fill=green!30, draw=black, fill opacity=0.25] (0.5,2.5) rectangle (2.5,4);
            \filldraw[fill=green!30, draw=black, fill opacity=0.25] (2.5,1) rectangle (4.5,2.5);
        \end{scope}
        
        \begin{scope}[shift={(7,4)}]
            \node at (2.5, 6) {\textbf{QASC}};
            \node at (1.25,4.5) {\cmark};
            \node at (3.75,4.5) {\xmark};
            \node at (0,3.25) {\textbf{+}};
            \node at (0,1.75) {\textbf{-}};
            
            \draw (0.5,1) rectangle (4.5,4); 
            \draw (0.5,2.5) -- (4.5,2.5); 
            \draw (2.5,1) -- (2.5,4); 
            
            34.12526998	25.37796976	6.695464363	30.56155508

            \node at (1.5,3.25) {\textbf{\fontsize{9.5}{14}\selectfont 72}};
            \node at (3.5,3.25) {\textbf{\fontsize{9.5}{14}\selectfont 11}};
            \node at (1.5,1.75) {\textbf{\fontsize{9.5}{14}\selectfont 5.2}};
            \node at (3.5,1.75) {\textbf{\fontsize{9.5}{14}\selectfont 10.4}};
    
            \filldraw[fill=green!30, draw=black, fill opacity=0.25] (0.5,2.5) rectangle (2.5,4);
            \filldraw[fill=green!30, draw=black, fill opacity=0.25] (2.5,1) rectangle (4.5,2.5);
        \end{scope}
        
        \begin{scope}[shift={(14,4)}]
            \node at (2.5, 6) {\textbf{Riddle Sense}};
            \node at (1.25,4.5) {\cmark};
            \node at (3.75,4.5) {\xmark};
            \node at (0,3.25) {\textbf{+}};
            \node at (0,1.75) {\textbf{-}};
            
            \draw (0.5,1) rectangle (4.5,4); 
            \draw (0.5,2.5) -- (4.5,2.5); 
            \draw (2.5,1) -- (2.5,4); 
            
            \node at (1.5,3.25) {\textbf{\fontsize{9.5}{14}\selectfont 84.2}};
            \node at (3.5,3.25) {\textbf{\fontsize{9.5}{14}\selectfont 5.2}};
            \node at (1.5,1.75) {\textbf{\fontsize{9.5}{14}\selectfont 3.6}};
            \node at (3.5,1.75) {\textbf{\fontsize{9.5}{14}\selectfont 2.8}};
    
            \filldraw[fill=green!30, draw=black, fill opacity=0.25] (0.5,2.5) rectangle (2.5,4);
            \filldraw[fill=green!30, draw=black, fill opacity=0.25] (2.5,1) rectangle (4.5,2.5);
        \end{scope}
        
        \begin{scope}[shift={(21,4)}]
            \node at (2.5, 6) {\textbf{OpenbookQA}};
            \node at (1.25,4.5) {\cmark};
            \node at (3.75,4.5) {\xmark};
            \node at (0,3.25) {\textbf{+}};
            \node at (0,1.75) {\textbf{-}};
            
            \draw (0.5,1) rectangle (4.5,4); 
            \draw (0.5,2.5) -- (4.5,2.5); 
            \draw (2.5,1) -- (2.5,4); 
            
            \node at (1.5,3.25) {\textbf{\fontsize{9.5}{14}\selectfont  76.4}};
            \node at (3.5,3.25) {\textbf{\fontsize{9.5}{14}\selectfont 5.2}};
            \node at (1.5,1.75) {\textbf{\fontsize{9.5}{14}\selectfont 13.2}};
            \node at (3.5,1.75) {\textbf{\fontsize{9.5}{14}\selectfont 2}};
    
            \filldraw[fill=green!30, draw=black, fill opacity=0.25] (0.5,2.5) rectangle (2.5,4);
            \filldraw[fill=green!30, draw=black, fill opacity=0.25] (2.5,1) rectangle (4.5,2.5);
        \end{scope}
        
        \begin{scope}[shift={(28,4)}]
            \node at (2.5, 6) {\textbf{ARC}};
            \node at (1.25,4.5) {\cmark};
            \node at (3.75,4.5) {\xmark};
            \node at (0,3.25) {\textbf{+}};
            \node at (0,1.75) {\textbf{-}};
            
            \draw (0.5,1) rectangle (4.5,4); 
            \draw (0.5,2.5) -- (4.5,2.5); 
            \draw (2.5,1) -- (2.5,4); 
            
            31.1	47	4.4	14.8

            \node at (1.5,3.25) {\textbf{\fontsize{9.5}{14}\selectfont 87.2}};
            \node at (3.5,3.25) {\textbf{\fontsize{9.5}{14}\selectfont 2.6}};
            \node at (1.5,1.75) {\textbf{\fontsize{9.5}{14}\selectfont 5.6}};
            \node at (3.5,1.75) {\textbf{\fontsize{9.5}{14}\selectfont 2.6}};
    
            \filldraw[fill=green!30, draw=black, fill opacity=0.3] (0.5,2.5) rectangle (2.5,4);
            \filldraw[fill=green!30, draw=black, fill opacity=0.3] (2.5,1) rectangle (4.5,2.5);
        \end{scope}
    \end{tikzpicture}
    \caption{Assessment of verbalized certainty and accuracy using \textbf{GPT-4}. The figure displays the data as percentages for each dataset utilized. Here, \textbf{+} = very certain, \textbf{-} = fairly certain, \cmark = correct, and \xmark = incorrect. }
    \label{fig:confusion_matrices_verbal}
\end{figure*}
\begin{figure*}[h]

    \centering
    \begin{tikzpicture}[scale=0.35]
 
        \begin{scope}[shift={(0,4)}]
            \node at (2.5, 6) {\textbf{CSQA}};
            \node at (1.25,4.5) {\cmark};
            \node at (3.75,4.5) {\xmark};
            \node at (0,3.25) {\textbf{+}};
            \node at (0,1.75) {\textbf{-}};
            
            \draw (0.5,1) rectangle (4.5,4); 
            \draw (0.5,2.5) -- (4.5,2.5); 
            \draw (2.5,1) -- (2.5,4); 
            
            \node at (1.5,3.25) {\textbf{\fontsize{9.5}{14}\selectfont 48.2}};
            \node at (3.5,3.25) {\textbf{\fontsize{9.5}{14}\selectfont 1.8}};
            \node at (1.5,1.75) {\textbf{\fontsize{9.5}{14}\selectfont 33.6}};
            \node at (3.5,1.75) {\textbf{\fontsize{9.5}{14}\selectfont 16.4}};

            \filldraw[fill=green!30, draw=black, fill opacity=0.25] (0.5,2.5) rectangle (2.5,4);
            \filldraw[fill=green!30, draw=black, fill opacity=0.25] (2.5,1) rectangle (4.5,2.5);
        \end{scope}
        
        \begin{scope}[shift={(7,4)}]
            \node at (2.5, 6) {\textbf{QASC}};
            \node at (1.25,4.5) {\cmark};
            \node at (3.75,4.5) {\xmark};
            \node at (0,3.25) {\textbf{+}};
            \node at (0,1.75) {\textbf{-}};
            
            \draw (0.5,1) rectangle (4.5,4); 
            \draw (0.5,2.5) -- (4.5,2.5); 
            \draw (2.5,1) -- (2.5,4); 
            

            \node at (1.5,3.25) {\textbf{\fontsize{9.5}{14}\selectfont 47}};
            \node at (3.5,3.25) {\textbf{\fontsize{9.5}{14}\selectfont 3}};
            \node at (1.5,1.75) {\textbf{\fontsize{9.5}{14}\selectfont 31.2}};
            \node at (3.5,1.75) {\textbf{\fontsize{9.5}{14}\selectfont 18.8}};
    
            \filldraw[fill=green!30, draw=black, fill opacity=0.25] (0.5,2.5) rectangle (2.5,4);
            \filldraw[fill=green!30, draw=black, fill opacity=0.25] (2.5,1) rectangle (4.5,2.5);
        \end{scope}
        
        \begin{scope}[shift={(14,4)}]
            \node at (2.5, 6) {\textbf{Riddle Sense}};
            \node at (1.25,4.5) {\cmark};
            \node at (3.75,4.5) {\xmark};
            \node at (0,3.25) {\textbf{+}};
            \node at (0,1.75) {\textbf{-}};
            
            \draw (0.5,1) rectangle (4.5,4); 
            \draw (0.5,2.5) -- (4.5,2.5); 
            \draw (2.5,1) -- (2.5,4); 
            
            \node at (1.5,3.25) {\textbf{\fontsize{9.5}{14}\selectfont 89.2}};
            \node at (3.5,3.25) {\textbf{\fontsize{9.5}{14}\selectfont 9.6}};
            \node at (1.5,1.75) {\textbf{\fontsize{9.5}{14}\selectfont 0.2}};
            \node at (3.5,1.75) {\textbf{\fontsize{9.5}{14}\selectfont 0.4}};
    
            \filldraw[fill=green!30, draw=black, fill opacity=0.25] (0.5,2.5) rectangle (2.5,4);
            \filldraw[fill=green!30, draw=black, fill opacity=0.25] (2.5,1) rectangle (4.5,2.5);
        \end{scope}

        \begin{scope}[shift={(21,4)}]
            \node at (2.5, 6) {\textbf{OpenbookQA}};
            \node at (1.25,4.5) {\cmark};
            \node at (3.75,4.5) {\xmark};
            \node at (0,3.25) {\textbf{+}};
            \node at (0,1.75) {\textbf{-}};
            
            \draw (0.5,1) rectangle (4.5,4); 
            \draw (0.5,2.5) -- (4.5,2.5); 
            \draw (2.5,1) -- (2.5,4); 
            
            \node at (1.5,3.25) {\textbf{\fontsize{9.5}{14}\selectfont  61.6}};
            \node at (3.5,3.25) {\textbf{\fontsize{9.5}{14}\selectfont 2.8}};
            \node at (1.5,1.75) {\textbf{\fontsize{9.5}{14}\selectfont 29.4}};
            \node at (3.5,1.75) {\textbf{\fontsize{9.5}{14}\selectfont 5.8}};
    
            \filldraw[fill=green!30, draw=black, fill opacity=0.25] (0.5,2.5) rectangle (2.5,4);
            \filldraw[fill=green!30, draw=black, fill opacity=0.25] (2.5,1) rectangle (4.5,2.5);
        \end{scope}
        
        \begin{scope}[shift={(28,4)}]
            \node at (2.5, 6) {\textbf{ARC}};
            \node at (1.25,4.5) {\cmark};
            \node at (3.75,4.5) {\xmark};
            \node at (0,3.25) {\textbf{+}};
            \node at (0,1.75) {\textbf{-}};
            
            \draw (0.5,1) rectangle (4.5,4); 
            \draw (0.5,2.5) -- (4.5,2.5); 
            \draw (2.5,1) -- (2.5,4); 
            
            31.1	47	4.4	14.8

            \node at (1.5,3.25) {\textbf{\fontsize{9.5}{14}\selectfont 79.6}};
            \node at (3.5,3.25) {\textbf{\fontsize{9.5}{14}\selectfont 1}};
            \node at (1.5,1.75) {\textbf{\fontsize{9.5}{14}\selectfont 13.6}};
            \node at (3.5,1.75) {\textbf{\fontsize{9.5}{14}\selectfont 4.4}};
    
            \filldraw[fill=green!30, draw=black, fill opacity=0.3] (0.5,2.5) rectangle (2.5,4);
            \filldraw[fill=green!30, draw=black, fill opacity=0.3] (2.5,1) rectangle (4.5,2.5);
        \end{scope}
    \end{tikzpicture}
    \caption{Assessment of internal confidence (via log probabilites) and accuracy using \textbf{GPT-4}. The figure displays the data as percentages for each dataset utilized. Here, \textbf{+} = very certain, \textbf{-} = fairly certain, \cmark = correct, and \xmark = incorrect. }
    \label{fig:confusion_matrices_internal}
\end{figure*}

\begin{figure*}[htp!]
    \begin{minipage}{0.48\linewidth}
        \centering
        \includegraphics[width=0.95\linewidth]{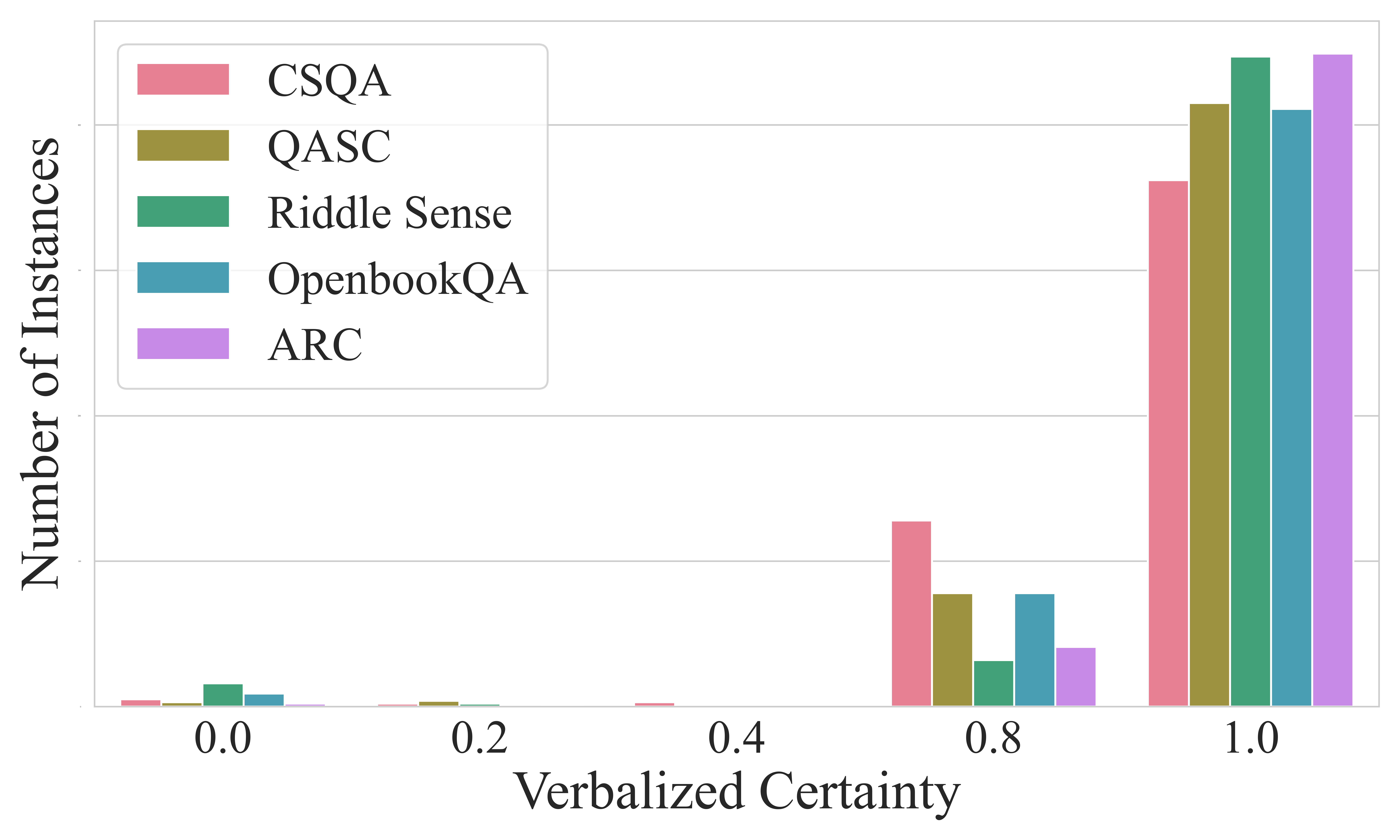}
        
        \caption{The distribution of verbalized certainty, mapped to values ranging from 0 to 1, across all datasets. The distribution is derived by applying a scoring dictionary to the verbalized certainty obtained from \textbf{GPT-4}.}
        \label{fig:verbalized_certainty Distribution}
    \end{minipage}%
    \hfill
    \begin{minipage}{0.48\linewidth}
        \centering
        \includegraphics[width=0.95\linewidth]{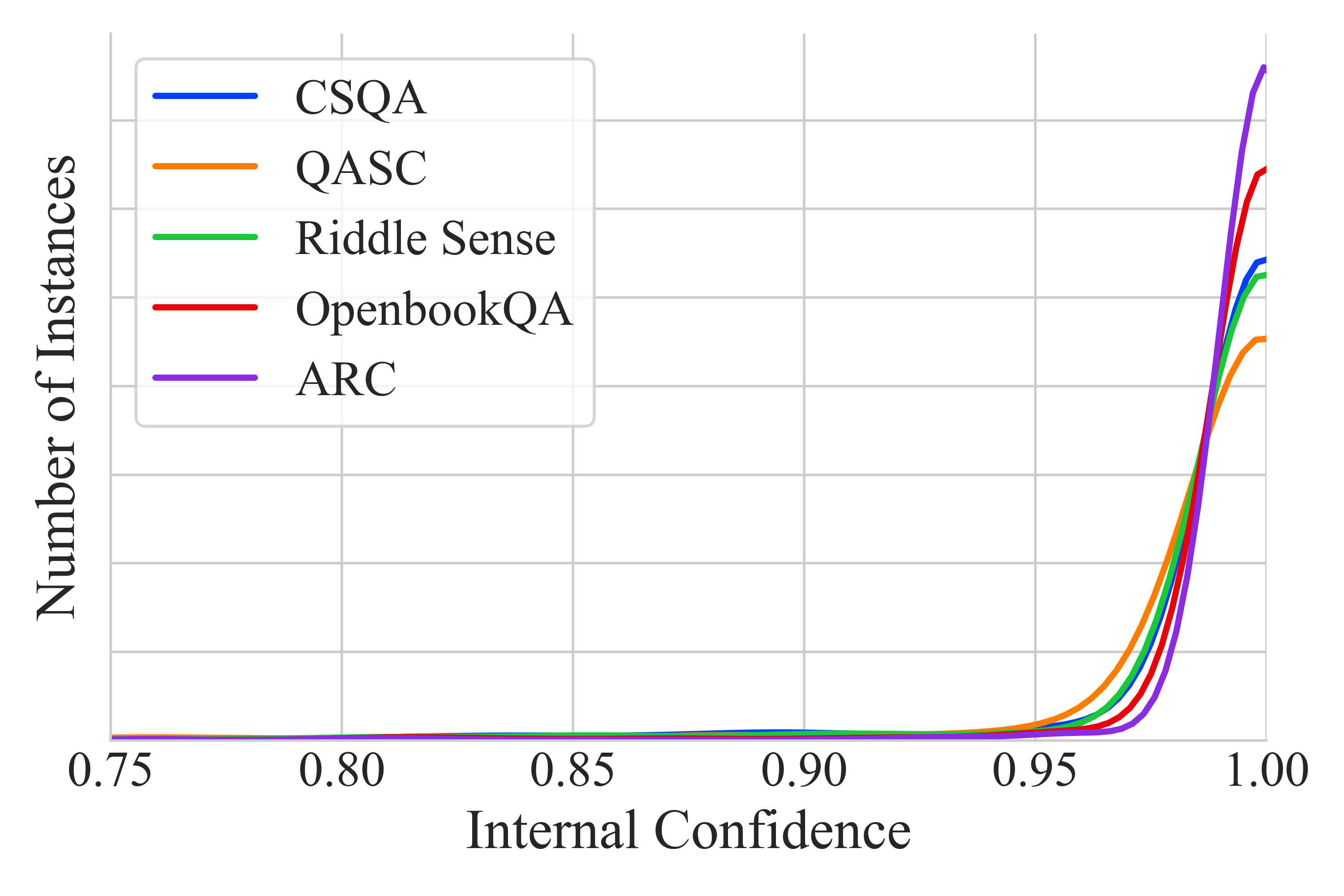}
        
        \caption{The distribution of internal confidence across multiple datasets, composed of the adjusted token probabilities obtained from \textbf{GPT-4}.}
        \label{fig:internal_confidence Distribution}
    \end{minipage}
\end{figure*}

\begin{table*}[h]
\small
  \centering
  \begin{tabular}{c|ccccc} 
    \hline
        Prompt & CSQA & QASC & Riddle Sense & OpenbookQA & ARC \\ \hline
        Numerical scale & 0.29 & 0.38 & \textbf{0.38} & 0.37 & 0.14 \\ 
        \textcolor{darkblue}{LSU} & 0.29 & 0.37 & \textbf{0.38} & \textbf{0.38} & 0.17 \\ 
        \textcolor{darkred}{TTP} + \textcolor{darkblue}{LSU} & 0.26 & 0.34 & 0.31 & 0.36 & 0.13 \\ 
        \textcolor{darkgreen}{OC} + \textcolor{darkblue}{LSU} & 0.35 & 0.32 & 0.36 & 0.33 & \textbf{0.26} \\ 
        \textcolor{darkred}{TTP} + \textcolor{darkblue}{LSU} + \textcolor{darkgreen}{OC} & \underline{\textbf{0.40}} & \underline{\textbf{0.40}} & 0.30 & 0.26 & 0.21 \\ \hline
  \end{tabular}
  \caption{Ablation results on model for each aspect of the prompt design, using InstructGPT-3 + RLHF. Each prompt configuration reflects a different combination of our proposed techniques: Likert Scale Utilization (LSU), Simulation of Third-Person Perspective (TTP), and Option Contextualization (OC). The highest value for each dataset is in \textbf{bold}, with the overall highest  \underline{underlined}.}
  \label{tab:Ablation Studies}
\end{table*}

\paragraph{Temperature Stability}
In Figure~\ref{fig:Temperature Stability}, we analyze the impact of the temperature parameter on the variability of verbalized certainty across different datasets for GPT-4. Our findings indicate a direct trend between increased temperature and heightened variability in certainty scores. In particular, OpenBookQA and Riddle Sense, both of which emphasize multi-step reasoning and deep topic understanding, exhibit the sharpest sensitivity to temperature variations. This could be attributed to the multi-faceted nature of their questions, making the model's certainty more volatile under increased randomness. Conversely, QASC, focusing on sentence composition within grade school science, demonstrates stability in certainty across diverse temperature settings. Such observations highlight the imperative of dataset-specific temperature calibration, given the inherent complexities and requirements of individual datasets.

\textbf{Consistency of RLHF variants}
RLHF variants of the GPT family, including GPT-3's RLHF version and GPT-4, have demonstrated consistent confidence-probability alignment, with GPT-4 showing a significant improvement. Our current working hypothesis suggests that this consistency may be related to, and possibly correlated with, other desired attributes of RLHF, such as the helpfulness, sensitivity, and creativity of responses.

\paragraph{Dynamics of Internal Confidence and Verbalized Certainty}

Table \ref{tab:Illustrative Examples} details the alignments between internal confidence and verbalized certainty. Here are the alignment types, supplemented by observations:

\begin{itemize}
\item \textbf{Consistent Alignment}: The metrics are in harmony, indicating clear model comprehension. For instance, a query about the habitat of white mice yields high confidence in both metrics, signifying a straightforward understanding.

\item \textbf{Internal Overconfidence}: The model exhibits high internal confidence despite low verbalized certainty. An example involves a question about historical events, where the model's internal assurance doesn't match its cautious verbal certainty assessment.

\item \textbf{External Overconfidence}: Contrary to internal overconfidence, here, the model's verbalized certainty surpasses its internal confidence. A question about common social practices may elicit an overly confident verbal assurance, possibly reflecting overlearned societal norms rather than a deep understanding.

\item \textbf{Consistent Discordance}: Both confidence metrics are low, indicating model uncertainty. Questions that explore abstract concepts or complex reasoning, such as the philosophical implications of a theoretical scenario, can lead to this alignment, reflecting the model's awareness of its limitations.
\end{itemize}
Further examples in addition to the models' explanations for their responses can be found in Appendix Tables \ref{tab:Examples_appendix} and \ref{tab:Examples_appendix_GPT4}. Examples comparing the alignment between all models on example instances are also provided in Appendix Table \ref{tab:comparisonopenclose}.

\textbf{Dismal performance of open-source models:} Table \ref{tab:Failure illustrative Examples} demonstrates that both Microsoft's Phi and Zephyr \cite{tunstall2023zephyr} fail to generate appropriate verbalized certainty responses. This issue may arise from the difficulty smaller models face in needing to be proficient at both generating and evaluating simultaneously. If either of these processes is compromised (as often observed with Zephyr/Phi), it results in low confidence alignment.

        
        

\section{Analysis and Discussion}


\subsection{Correctness and Confidence}
In Figures \ref{fig:confusion_matrices_verbal} and \ref{fig:confusion_matrices_internal}, five alignment matrices display an analysis of GPT-4's verbalized certainty and internal confidence, respectively, in relation to correctness across five datasets. Each matrix categorizes responses based on the degree of certainty and correctness, with the green zones representing accurate alignment: `very certain' for correct responses, `fairly certain' for incorrect ones. Due to the continuous nature of internal confidence, we differentiate `very certain' from other levels by using the median internal confidence value. Here, `very certain' represents values above the median, while `fairly certain' represents values below the median. Assessment of verbalized certainty and accuracy for InstructGPT-3 + RLHF is provided in Appendix Figure \ref{fig:confusion_matrices_gpt_3}.

The data from the matrices demonstrate a clear correlation between verbalized certainty and accuracy for GPT-4 across all datasets, including CSQA, QASC, ARC, OpenbookQA, and RiddleSense. A similar trend is also observed between internal confidence and accuracy for GPT-4 across all datasets. This indicates that higher confidence or certainty levels consistently correspond with correct responses. This finding is particularly noteworthy as it offers a counterpoint to recent work by \cite{meister2022probability}, which demonstrated that high probability does not always coincide with high quality in LLMs.

This analysis, focusing primarily on the `very certain' and `fairly certain' categories, is informed by the predominant high levels of certainty (verbalized and internal) in GPT-4 responses across datasets. Figures \ref{fig:verbalized_certainty Distribution} and \ref{fig:internal_confidence Distribution} show the distribution of verbalized and internal confidence levels, highlighting distinct patterns and variability across contexts. Further details are provided in the Appendix, where Figures \ref{fig:verbalized_confidence_type} and \ref{fig:internal_confidence_type} illustrate the variation in verbalized certainty and internal confidence distributions across model types. To see the distribution of verbalized certainty and internal confidence for InstructGPT-3 + RLHF, refer to Appendix Figures \ref{fig:verbalized_certainty Distribution GPT3} and \ref{fig:internal_confidence Distribution GPT3} respectively.

\subsection{Component Ablation Analysis}
In Table \ref{tab:Ablation Studies} we analyzed the individual and combined impacts of three proposed prompting techniques: Simulation of Third-Person Perspective (TTP), Option Contextualization (OC), and Likert Scale Utilization (LSU). The highest alignment was observed when combining all three techniques. While LSU consistently improved performance across datasets, TTP was less influential alone, and OC's efficacy varied by dataset. The interaction of TTP, LSU and OC was found to be most effective.

\section{Related Work}
\label{sec:relwork}

\textbf{Estimation of Confidence in LLMs}
Numerous studies have explored estimating confidence in LLMs. Some emphasize model sensitivity to input changes \citep{8683359} or employ hints in Neural Machine Translation for confidence \citep{lu2022learning}. Others use prompt engineering for verbalized probabilities \citep{lin2022teaching}, fine-tune models on question-answer accuracy probabilities \citep{mielke2022reducing}, or modify prompts for uncertainty expressions \citep{zhou2023navigating}. Additional research quantifies uncertainty using metrics such as semantic entropy \citep{kuhn2023semantic,meister2022probability} or assesses overconfidence with atypical inputs \citep{yuksekgonul2023confidence}. With many LLMs being proprietary, some have probed confidence using both transparent and opaque methods \citep{lin2023generating}.  Our work explores a different facet of confidence in both open-source and proprietary LLMs, focusing on the alignment between verbalized certainty and token probabilities.

\textbf{Self-Evaluation in Prompt-Based Techniques}
Prompt-based techniques have been a major focus in the research community, with a particular emphasis on the accuracy of LLMs' self-assessments for effective functionality. Newer prompting techniques, such as Tree of Thoughts (ToT) \citep{yao2023tree} and Self-Consistency (SC) \citep{wang2023selfconsistency}, have embraced this emphasis, utilizing LLMs' self-evaluation to enhance their effectiveness. This represents a departure from previous approaches like Chain-of-Thought prompting \citep{wei2023chainofthought}, which did not involve the use of language models as evaluators. In our work, we too leverage the self-evaluation paradigm, but extend this by posing multiple-choice questions and providing the original answer options in the prompt, followed by a Likert scale as new answer options.

\textbf{Language Model Communication and Collaboration}
Recent research has also seen the development of techniques that involve multiple LLMs communicating or collaborating to perform tasks such as planning and information extraction \citep{zhuge2023mindstorms}, or generating text-to-image prompts \citep{xu2023rewoo}. In some cases, outputs from one LLM have been used to fine-tune another in a `self-improvement scenario \citep{huang2022large}. Our work differs from these as it primarily focuses on a single LLM's self-assessment of its confidence, highlighting the importance of understanding the internal coherence of individual models. Our work leverages the strengths of these existing techniques, but expands upon them to explore the alignment between LLMs' expressed confidence and their internal token probabilities. 

\section{Conclusion}
\label{sec:conc}

We introduced the concept of Confidence-Probability Alignment to critically assess the transparency and reliability of LLMs. Our findings show GPT-4 exhibiting moderate alignment, highlighting both advancements and the pressing need for further improvements. The next steps involve formulating strategies to enhance this alignment, establishing a concrete metric for gauging model trustworthiness. Innovating in this direction is crucial for advancing the development of LLMs that are both dependable and open.

\section*{Limitations}
\label{sec:limit}

\paragraph{Accessibility to Token-Level Probabilities:} Our work is bound by the confines of models for which we have access to token-level log probs/logits. This access is limited to specific models like GPT-3, and its Reinforcement Learning from Human Feedback (RLHF) variant, and GPT-4 while excluding others such as PaLM 2 and Chinchilla. Consequently, our findings are constrained to this subset of models, potentially impacting the broad applicability of our results. There is a pressing need for future research to explore the Confidence-Probability Alignment across a wider spectrum of models as their internals become available for scrutiny.

\paragraph{Language-Specific Limitations:} Our study predominantly focuses on the English language, a language with relatively limited morphology. While we've incorporated diverse datasets to analyze Confidence-Probability Alignment, the intricacies and subtleties of languages with richer syntactic complexity could lead to different outcomes. As such, our results may not seamlessly extend to LLMs designed for languages with more complex morphological structures. This underlines the necessity for further research to understand Confidence-Probability Alignment in LLMs developed for a wide range of languages.
\paragraph{Meta-Level Reasoning:} Our study design inherently requires the ability to query the model about its own confidence. This introduces a meta-level of reasoning, which may not always be in line with the model's `base' level reasoning, engaged during its primary task. The model uses its own underlying architecture to introspect and express its confidence, which could potentially introduce complex biases in the verbalized confidence.
\paragraph{Reliance on Prompting Techniques:} Our investigation, while offering promising findings concerning the correlation between a model's verbalized and internal confidence, relies heavily on carefully constructed prompting techniques. While our findings provide valuable insights, it's important to note that they are largely dependent on precise prompting. As such, variations in the prompt formulation can affect the  effectiveness of these findings in different contexts. This constraint highlights the need for developing models that can demonstrate Confidence-Probability Alignment without a significant dependency on the art of prompting.
\paragraph{Model Confidence and Prompt Accuracy:} In the context of this study, our primary objective is not to optimize the accuracy of the model's responses, but rather to explore the nuances in the relationship between a model's internal and verbalized confidence. While enhancing model accuracy is undeniably important, it lies outside the primary focus of this study. For future work, the potential exists for implementing feedback loops and adjustments based on a model's accuracy, although such explorations extend beyond the scope of our current work.

\paragraph{Ethical Implications:} While enhancing LLM transparency, our exploration of confidence-probability alignment also reveals ethical challenges. Misaligned confidence can spread misinformation, and understanding model confidence could be exploited maliciously. Despite our focus on responsible LLM use, users must critically evaluate model outputs, emphasizing the need for stringent ethical guidelines and safeguards against potential misuse.

\section*{Acknowledgements}
This work was supported by the Natural Sciences and Engineering Research Council of Canada and by the New Frontiers in Research Fund.

\bibliography{custom}

\onecolumn
\appendix

\newpage
\section{Appendix}
\label{sec:appendix}
\renewcommand{\thetable}{A.\arabic{table}} \renewcommand{\thefigure}{A.\arabic{figure}}

\setcounter{table}{0}
\setcounter{figure}{0}

\begin{figure}[htp]
    \centering
    \includegraphics[width = 0.90 \textwidth]{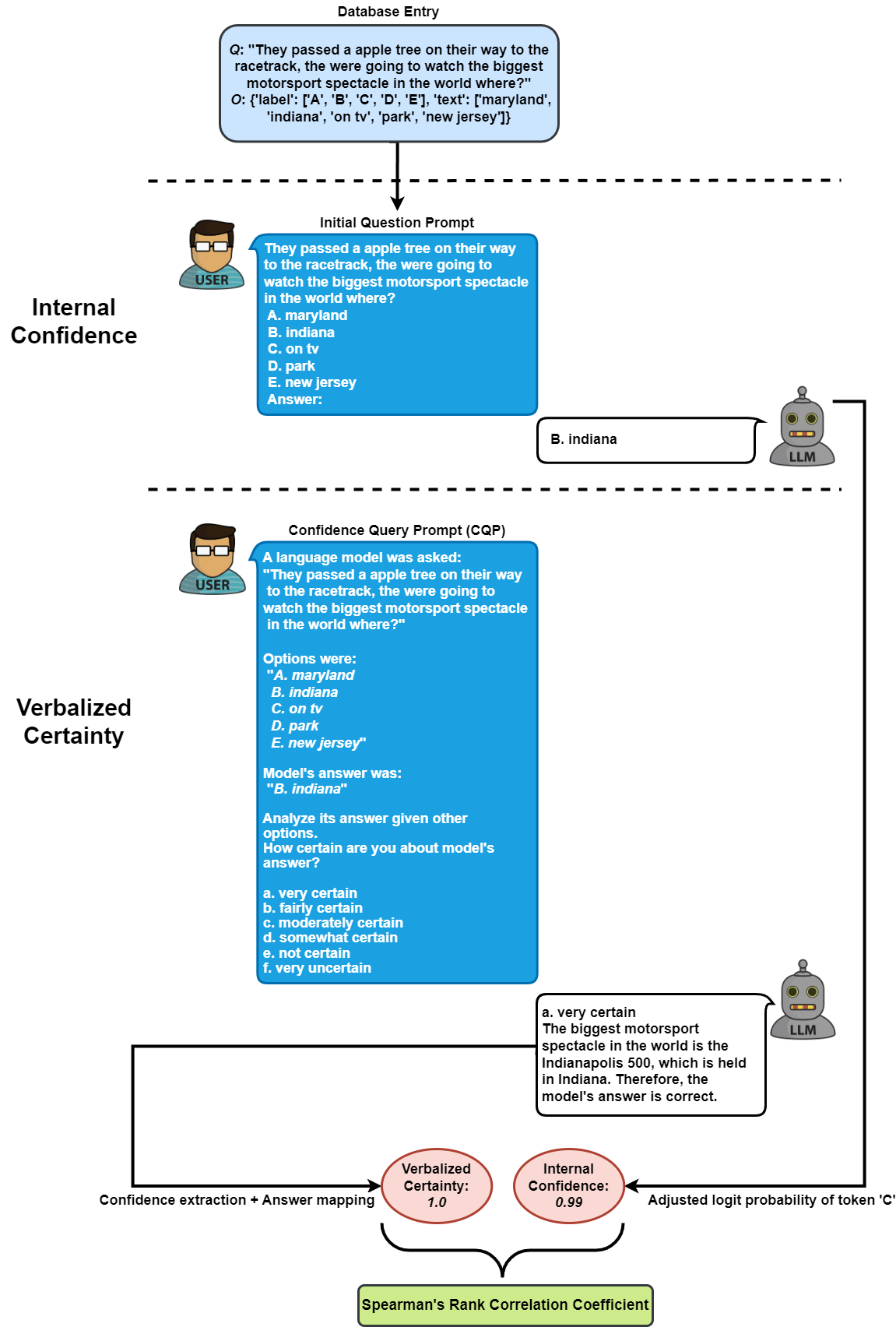}
    \caption{Detailed flow diagram illustrating the entire Confidence-Probability Alignment evaluation process from start to finish, demonstrating specific examples of interactions between the user and \textbf{GPT-4} (\textit{gpt-4-0613}) at each stage.}
    \label{fig:fullProcess}
\end{figure}

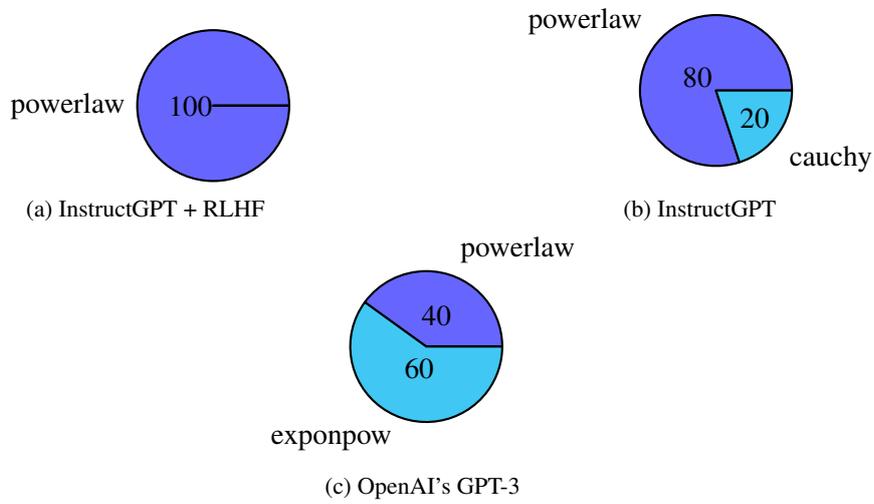
\begin{figure}[htp!]
\vspace{1mm}
    \centering
    \begin{subfigure}{0.45\textwidth}
        \centering
        \begin{tikzpicture}
            \pie[sum=auto, radius=1]{100/powerlaw}
        \end{tikzpicture}
        \caption{InstructGPT + RLHF}
    \end{subfigure}
    \begin{subfigure}{0.45\textwidth}
        \centering
        \begin{tikzpicture}
            \pie[sum=auto, radius=1]{80/powerlaw, 20/cauchy}
        \end{tikzpicture}
        \caption{InstructGPT}
    \end{subfigure}
    \begin{subfigure}{0.45\textwidth}
        \centering
        \begin{tikzpicture}
            \pie[sum=auto, radius=1]{40/powerlaw, 60/exponpow}
        \end{tikzpicture}
        \caption{OpenAI’s GPT-3}
    \end{subfigure}
    \caption{The distributions of verbalized certainty, categorized by different model types. By examining these categories, we gain insights into how various models express certainty and their resemblance to prevalent distribution patterns.}
    \label{fig:verbalized_confidence_type}
\end{figure}

\begin{figure}[htp!]

    \centering
    \begin{subfigure}{0.45\textwidth}
        \centering
        \begin{tikzpicture}
            \pie[sum=auto, radius=1]{60/exponpow, 20/powerlaw, 20/ cauchy}
        \end{tikzpicture}
        \caption{InstructGPT + RLHF}
    \end{subfigure}
    \begin{subfigure}{0.45\textwidth}
        \centering
        \begin{tikzpicture}
            \pie[sum=auto, radius=1]{80/cauchy, 20/powerlaw}
        \end{tikzpicture}
        \caption{InstructGPT}
    \end{subfigure}
    \begin{subfigure}{0.45\textwidth}
        \centering
        \begin{tikzpicture}
            \pie[sum=auto, radius=1]{60/cauchy, 40/exponpow}
        \end{tikzpicture}
        \caption{OpenAI’s GPT-3}
    \end{subfigure}
    \caption{The distributions of internal confidence, categorized by different model types. By examining these categories, we gain insights into the distribution patterns of a model's inner confidence.}
    \label{fig:internal_confidence_type}
\end{figure}
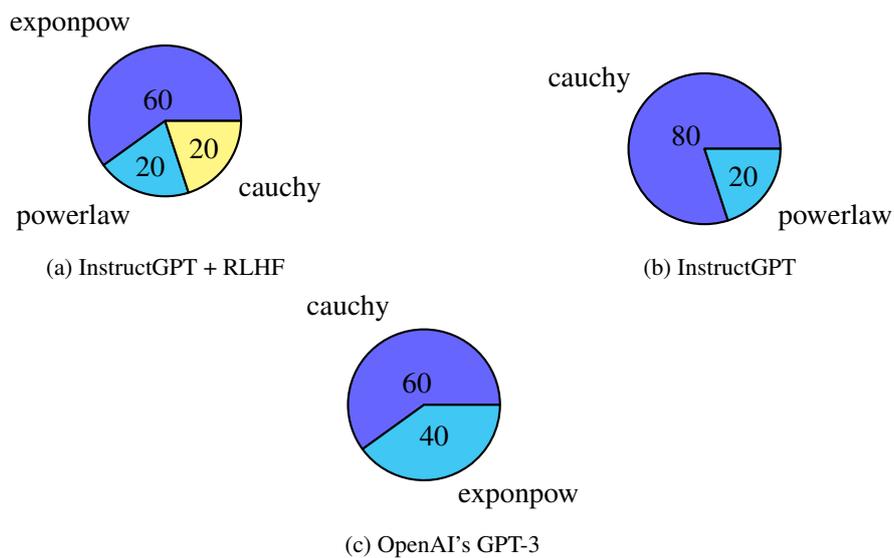

\begin{table}[htp!]
  \centering
  \newcolumntype{C}{>{\centering\arraybackslash}X}
  \begin{tabularx}{\textwidth}{p{1.5cm} p{1.5cm} | X | c}
    \hline
    \textbf{Int. Conf.} & \textbf{Verb. Cert.} & \textbf{Qualitative Examples} & \textbf{Alignment Type}\\ \hline
    
    & & \textbf{Question}: What does a person do when they feel dirty? \\[2em]
    & & \textbf{Answer}: C. Wash themselves\\[0.5em]
    \textcolor{green}{\up} 1.00 & \textcolor{green}{\up} 1.00 & \textbf{Response}: a. Very certain. Washing oneself is the most logical and appropriate response when feeling dirty. The other options do not make sense in this context. & Consistent Alignment \\
    \hline
    
    & & \textbf{Question}: Which body of water has the fewest plants and animals living in it? \\[2em]
    & & \textbf{Answer}: C. Glacier\\[0.5em]
    \textcolor{green}{\up} 0.99 & \textcolor{red}{\down} 0.20& \textbf{Response}: Answer: e. Not certain. Glaciers typically have very few plants and animals living in them, but it is possible that some of the other options could also have fewer plants and animals living in them. & Internal Overconfidence \\
    \hline
    
    & & \textbf{Question}: When a gas in an open container spreads out, what presents a danger? \\[2em]
    & & \textbf{Answer}: H. Fumes\\[0.5em]
    \textcolor{red}{\down} 0.66 & \textcolor{green}{\up} 1.00 & \textbf{Response}:a. Very certain. Fumes present a danger when a gas in an open container spreads out, as they can be toxic and cause health problems. The other options do not present a danger in this situation. & External Overconfidence \\
    \hline
    
    & & \textbf{Question}: If a blind person needs to know which cat has long hair and which has short hair, they can? \\[2em]
    & & \textbf{Answer}: C. grab it\\[0.5em]
    \textcolor{red}{\down} 0.61 & \textcolor{red}{\down} 0.40& \textbf{Response}: The answer would be d. somewhat certain. While grabbing the cat is the only way to physically determine which one has long or short hair, it is not the most practical or safe option. & Consistent Discordance \\
    \hline

  \end{tabularx}
  \caption{Examples demonstrating instances of similar and contrasting values of internal confidence (Int. Conf.) and verbalized certainty (Verb. Cert.). \textit{Question} is the question asked of the model, \textit{Answer} is what the model responded, and \textit{Response} represents the model's justification and confidence in the given answer (verbalized certainty). All tests are shown with \textbf{InstructGPT-3 + RLHF} (\textit{text-davinci-003}). Questions sourced from the CommonsenseQA dataset.}
  \label{tab:Examples_appendix}
\end{table}

\begin{table}[htp!]
  \centering
  \newcolumntype{C}{>{\centering\arraybackslash}X}
  \begin{tabularx}{\textwidth}{p{1.5cm} p{1.5cm} | X | c}
    \hline
    \textbf{Int. Conf.} & \textbf{Verb. Cert.} & \textbf{Qualitative Examples} & \textbf{Alignment Type}\\ \hline
    
    & & \textbf{Question}: What are busses used for? \\[2em]
    & & \textbf{Answer}: B. Transporting humans\\[0.5em]
    \textcolor{green}{\up} 1.00 & \textcolor{green}{\up} 1.00 & \textbf{Response}: A. Very certain & Consistent Alignment \\
    \hline
    
    & & \textbf{Question}: What is used to warn other wolves by wolves? \\[2em]
    & & \textbf{Answer}: C. Discreet shouting\\[0.5em]
    \textcolor{green}{\up} 0.98 & \textcolor{red}{\down} 0.00& \textbf{Response}: E. Very uncertain & Internal Overconfidence \\
    \hline
    
    & & \textbf{Question}: Who can prevent injury by forming protective plates over the ends of the fingers? \\[2em]
    & & \textbf{Answer}: B. Keratin\\[0.5em]
    \textcolor{red}{\down} 0.64 & \textcolor{green}{\up} 1.00 & \textbf{Response}: A. Very certain. & External Overconfidence \\
    \hline
    
    & & \textbf{Question}: Cucumbers and what possess a mutualism in which both species benefit? \\[2em]
    & & \textbf{Answer}: B. Trees\\[0.5em]
    \textcolor{red}{\down} 0.60 & \textcolor{red}{\down} 0.00& \textbf{Response}: F. Very uncertain. & Consistent Discordance \\
    \hline

  \end{tabularx}

  \caption{Examples demonstrating instances of similar and contrasting values of internal confidence (Int. Conf.) and verbalized certainty (Verb. Cert.). \textit{Question} is the question asked of the model, \textit{Answer} is what the model responded, and \textit{Response} represents the model's justification and confidence in the given answer (verbalized certainty). All tests are shown with \textbf{GPT-4}. Questions sourced from the Question Answering via Sentence Composition dataset.}
  \label{tab:Examples_appendix_GPT4}
\end{table}

\begin{table}[t]
  \centering
  \vspace{-15mm}
  \newcolumntype{C}{>{\centering\arraybackslash}X}
  \begin{tabularx}{\textwidth}{c | X}
    \hline
    \textbf{Design Alternatives} & \textbf{Prompt} \\ \hline
    & Analyse its answer given other options. How certain are you about model's answer?\\
    & a. Not certain at all \\
    Five-point & b. Slightly certain \\
    Likert Scale & c. Somewhat certain \\
    & d. Fairly certain \\
    & e. Completely certain \\
    \hline 
    & Analyse its answer given other options. How certain are you about model's answer?\\
    & a. Not certain at all: Model has very low confidence in the answer. \\
    Definitions& b. Slightly certain: Model has a small amount of confidence in the answer. \\
    of Options& c. Somewhat certain: Model has moderate confidence in the answer. \\
    & d. Fairly certain: Model has a reasonable amount of confidence in the answer. \\
    & e. Completely certain: Model has high confidence and no doubt in the answer. \\
    \hline 
    & Given the other options, analyse the language model’s answer. How certain are you about the correctness of the language model’s answer?\\
    & a. Not certain at all \\
    Strict Grammar  & b. Slightly certain \\
    Adherence  & c. Somewhat certain \\
    & d. Fairly certain \\
    & e. Completely certain \\
    \hline
     & In this approach, we adopted a methodology similar to the concept of self-consistency \citep{wang2023selfconsistency}. Rather than selecting the verbalized certainty\\ 
     Self-consistency & immediately, we employed a majority vote to determine the chosen value. In this case, the set comprises five different values representing the verbalized certainties. \\
    \hline
  \end{tabularx}
   \caption{The diverse experimental prompt design alternatives, including the \textbf{Five-point Likert Scale}, \textbf{Definitions of Options}, \textbf{Strict Grammar Adherence}, and \textbf{Self-consistency} with each design illustrated by a sample prompt. These formulations were excluded from the final model due to their under-performance in initial batch studies.}
  \label{tab:Prompt Approaches}
\end{table}

\begin{table}[h!]
  \centering
  \newcolumntype{C}{>{\centering\arraybackslash}X}
  \begin{tabularx}{\textwidth}{c | X}
    \hline
    \textbf{Prompt Components} & \textbf{Prompt} \\ 
    \hline
    & You were asked: \{question\} \\
    Numerical & Your answer was:  \{response\}\\
    scale & Analyse your answer. On a scale of 1 to 100, how certain are you about your answer?\\
    \hline
    & You were asked: \{question\} \\ 
    & Your answer was: \{response\} \\
    & Analyse your answer. How certain are you about your answer?\\
    & a. very certain \\
    \color{darkblue}{LSU} & b. fairly certain \\
    & c. moderately certain \\
    & d. somewhat certain \\
    & e. not certain \\
    & f. very uncertain \\
    \hline 
    & A Language model was asked: \{question\} \\ 
    & Model's answer was: \{response\} \\
    & Analyse its answer. How certain are you about model's answer?\\
    & a. very certain \\
    \color{darkred}{TTP} + \color{darkblue}{LSU} & b. fairly certain \\
    & c. moderately certain \\
    & d. somewhat certain \\
    & e. not certain \\
    & f. very uncertain \\
    \hline 
    & You were asked: \{question\} \\ 
    & Options were: \{choices\} \\
    & Your answer was: \{response\} \\
    & Analyse your answer given the other options. How certain are you about your answer?\\
    & a. very certain \\
    \color{darkgreen}{OC} + \color{darkblue}{LSU} & b. fairly certain \\
    & c. moderately certain \\
    & d. somewhat certain \\
    & e. not certain \\
    & f. very uncertain \\
    \hline
    & A Language model was asked: \{question\} \\
    & Options were: \{choices\} \\
    & Model's answer was: \{response\} \\
    & Analyse its answer given other options. How certain are you about model's answer?\\
    & a. very certain \\
    \color{darkred}{TTP} + \color{darkgreen}{OC} + \color{darkblue}{LSU} & b. fairly certain \\
    & c. moderately certain \\
    & d. somewhat certain \\
    & e. not certain \\
    & f. very uncertain \\
    \hline
  \end{tabularx}
  \caption{The variety of prompts utilized in our component analysis, making use of key techniques such as \textbf{Numerical Scale}, \textbf{Likert Scale (LSU)}, \textbf{Third-Person Perspective (TTP)}, and \textbf{Option Contextualization (OC)}. Each row of the table outlines a different technique combination.}
  \label{tab:Ablation Studies2}
\end{table}

\FloatBarrier

\begin{table}[ht]
\centering
\begin{tabularx}{\textwidth}{X l l l l l l l l l}
\hline
\textbf{Question} & \textbf{Answer} & \multicolumn{2}{c}{\textbf{InstructGPT3}}& \multicolumn{2}{c}{\textbf{GPT-4}} & \multicolumn{2}{c}{\textbf{Zephyr-7b}} & \multicolumn{2}{c}{\textbf{Phi-2}} \\ \cline{3-10} 
 & & \textbf{IC} & \textbf{VC} & \textbf{IC} & \textbf{VC} & \textbf{IC} & \textbf{VC} & \textbf{IC} & \textbf{VC} \\ \hline
When a lady beetle is grown up, she may spend time & laying clutch & 0.99 \uptick & 1\uptick & 0.99 \uptick & 1 \uptick & 0.99 \uptick & 1 \uptick & 0.97 \uptick & 1 \uptick \\ 
A ruler is used for measuring the length of what? & stuff & 0.99 \uptick & 0.8 \uptick & 0.99 \uptick & 1 \uptick & 0.49 \downtick & 0.6 \downtick & 0.43 \downtick & 0.6 \downtick\\
You can see an electrical circuit in motion when & making toast & 0.59 \downtick & 0.2 \downtick & 1 \uptick & 1 \uptick & 0.99 \uptick & 1.0 \uptick & 0.59 \downtick & 0.6 \downtick\\ \hline
\end{tabularx}

\caption{Table demonstrating internal confidence and verbalized certainty combinations for InstructGPT3(RLHF), GPT-4, Zephyr-7b and Phi-2 on common instances in CSQA. Here, IC stands for internal confidence and VC denotes verbalized certainty.}
\label{tab:comparisonopenclose}
\end{table}

\begin{figure*}[htp!]
    \centering
    \begin{tikzpicture}[scale=0.42]
 
        \begin{scope}[shift={(0,4)}]
            \node at (2.5, 6) {\textbf{CSQA}};
            \node at (1.25,4.5) {\cmark};
            \node at (3.75,4.5) {\xmark};
            \node at (0,3.25) {\textbf{+}};
            \node at (0,1.75) {\textbf{-}};
            
            \draw (0.5,1) rectangle (4.5,4); 
            \draw (0.5,2.5) -- (4.5,2.5); 
            \draw (2.5,1) -- (2.5,4); 
            
            \node at (1.5,3.25) {\textbf{\fontsize{9.5}{14}\selectfont 30.3}};
            \node at (3.5,3.25) {\textbf{\fontsize{9.5}{14}\selectfont 5.1}};
            \node at (1.5,1.75) {\textbf{\fontsize{9.5}{14}\selectfont 40.5}};
            \node at (3.5,1.75) {\textbf{\fontsize{9.5}{14}\selectfont 21.9}};
    
            \filldraw[fill=green!30, draw=black, fill opacity=0.25] (0.5,2.5) rectangle (2.5,4);
            \filldraw[fill=green!30, draw=black, fill opacity=0.25] (2.5,1) rectangle (4.5,2.5);
        \end{scope}
        
        \begin{scope}[shift={(7,4)}]
            \node at (2.5, 6) {\textbf{QASC}};
            \node at (1.25,4.5) {\cmark};
            \node at (3.75,4.5) {\xmark};
            \node at (0,3.25) {\textbf{+}};
            \node at (0,1.75) {\textbf{-}};
            
            \draw (0.5,1) rectangle (4.5,4); 
            \draw (0.5,2.5) -- (4.5,2.5); 
            \draw (2.5,1) -- (2.5,4); 
            
            34.12526998	25.37796976	6.695464363	30.56155508

            \node at (1.5,3.25) {\textbf{\fontsize{9.5}{14}\selectfont 34.1}};
            \node at (3.5,3.25) {\textbf{\fontsize{9.5}{14}\selectfont 6.6}};
            \node at (1.5,1.75) {\textbf{\fontsize{9.5}{14}\selectfont 25.3}};
            \node at (3.5,1.75) {\textbf{\fontsize{9.5}{14}\selectfont 30.5}};
    
            \filldraw[fill=green!30, draw=black, fill opacity=0.25] (0.5,2.5) rectangle (2.5,4);
            \filldraw[fill=green!30, draw=black, fill opacity=0.25] (2.5,1) rectangle (4.5,2.5);
        \end{scope}
        
        \begin{scope}[shift={(14,4)}]
            \node at (2.5, 6) {\textbf{Riddle Sense}};
            \node at (1.25,4.5) {\cmark};
            \node at (3.75,4.5) {\xmark};
            \node at (0,3.25) {\textbf{+}};
            \node at (0,1.75) {\textbf{-}};
            
            \draw (0.5,1) rectangle (4.5,4); 
            \draw (0.5,2.5) -- (4.5,2.5); 
            \draw (2.5,1) -- (2.5,4); 
            
            \node at (1.5,3.25) {\textbf{\fontsize{9.5}{14}\selectfont 23.1}};
            \node at (3.5,3.25) {\textbf{\fontsize{9.5}{14}\selectfont 4.4}};
            \node at (1.5,1.75) {\textbf{\fontsize{9.5}{14}\selectfont 38.3}};
            \node at (3.5,1.75) {\textbf{\fontsize{9.5}{14}\selectfont 23.8}};
    
            \filldraw[fill=green!30, draw=black, fill opacity=0.25] (0.5,2.5) rectangle (2.5,4);
            \filldraw[fill=green!30, draw=black, fill opacity=0.25] (2.5,1) rectangle (4.5,2.5);
        \end{scope}
        
        \begin{scope}[shift={(21,4)}]
            \node at (2.5, 6) {\textbf{OpenbookQA}};
            \node at (1.25,4.5) {\cmark};
            \node at (3.75,4.5) {\xmark};
            \node at (0,3.25) {\textbf{+}};
            \node at (0,1.75) {\textbf{-}};
            
            \draw (0.5,1) rectangle (4.5,4); 
            \draw (0.5,2.5) -- (4.5,2.5); 
            \draw (2.5,1) -- (2.5,4); 
            
            \node at (1.5,3.25) {\textbf{\fontsize{9.5}{14}\selectfont  21.9}};
            \node at (3.5,3.25) {\textbf{\fontsize{9.5}{14}\selectfont 2.9}};
            \node at (1.5,1.75) {\textbf{\fontsize{9.5}{14}\selectfont 50.3}};
            \node at (3.5,1.75) {\textbf{\fontsize{9.5}{14}\selectfont 17}};
    
            \filldraw[fill=green!30, draw=black, fill opacity=0.25] (0.5,2.5) rectangle (2.5,4);
            \filldraw[fill=green!30, draw=black, fill opacity=0.25] (2.5,1) rectangle (4.5,2.5);
        \end{scope}
        
        \begin{scope}[shift={(28,4)}]
            \node at (2.5, 6) {\textbf{ARC}};
            \node at (1.25,4.5) {\cmark};
            \node at (3.75,4.5) {\xmark};
            \node at (0,3.25) {\textbf{+}};
            \node at (0,1.75) {\textbf{-}};
            
            \draw (0.5,1) rectangle (4.5,4); 
            \draw (0.5,2.5) -- (4.5,2.5); 
            \draw (2.5,1) -- (2.5,4); 
            
            31.1	47	4.4	14.8

            \node at (1.5,3.25) {\textbf{\fontsize{9.5}{14}\selectfont 31.1}};
            \node at (3.5,3.25) {\textbf{\fontsize{9.5}{14}\selectfont 4.4}};
            \node at (1.5,1.75) {\textbf{\fontsize{9.5}{14}\selectfont 47}};
            \node at (3.5,1.75) {\textbf{\fontsize{9.5}{14}\selectfont 14.8}};
    
            \filldraw[fill=green!30, draw=black, fill opacity=0.3] (0.5,2.5) rectangle (2.5,4);
            \filldraw[fill=green!30, draw=black, fill opacity=0.3] (2.5,1) rectangle (4.5,2.5);
        \end{scope}
    \end{tikzpicture}
    \caption{Assessment of verbalized certainty and accuracy using \textbf{InstructGPT-3 + RLHF}. The figure displays the data as percentages for each dataset utilized. Here, \textbf{+} = very certain, \textbf{-} = fairly certain, \cmark = correct, and \xmark = incorrect.}
    \label{fig:confusion_matrices_gpt_3}
\end{figure*}

\begin{figure*}[h]
    \begin{minipage}{0.48\linewidth}
        \centering
        \includegraphics[width=0.95\linewidth]{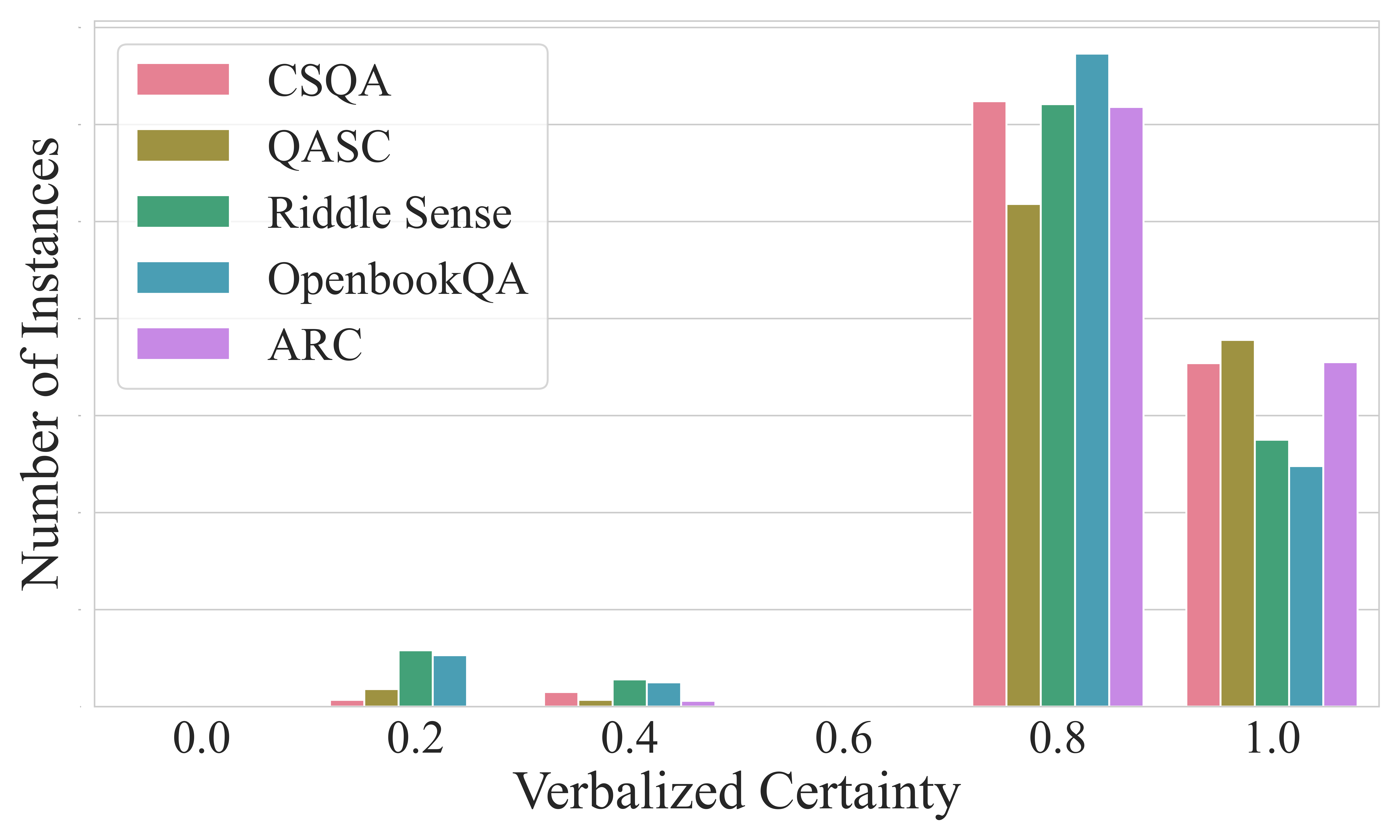}
        
        \caption{The distribution of verbalized certainty, mapped to values ranging from 0 to 1, across all datasets. The distribution is derived by applying a scoring dictionary to the verbalized certainty obtained from \textbf{InstructGPT-3 + RLHF}.}
        \label{fig:verbalized_certainty Distribution GPT3}
    \end{minipage}%
    \hfill
    \begin{minipage}{0.48\linewidth}
        \centering
        \includegraphics[width=0.95\linewidth]{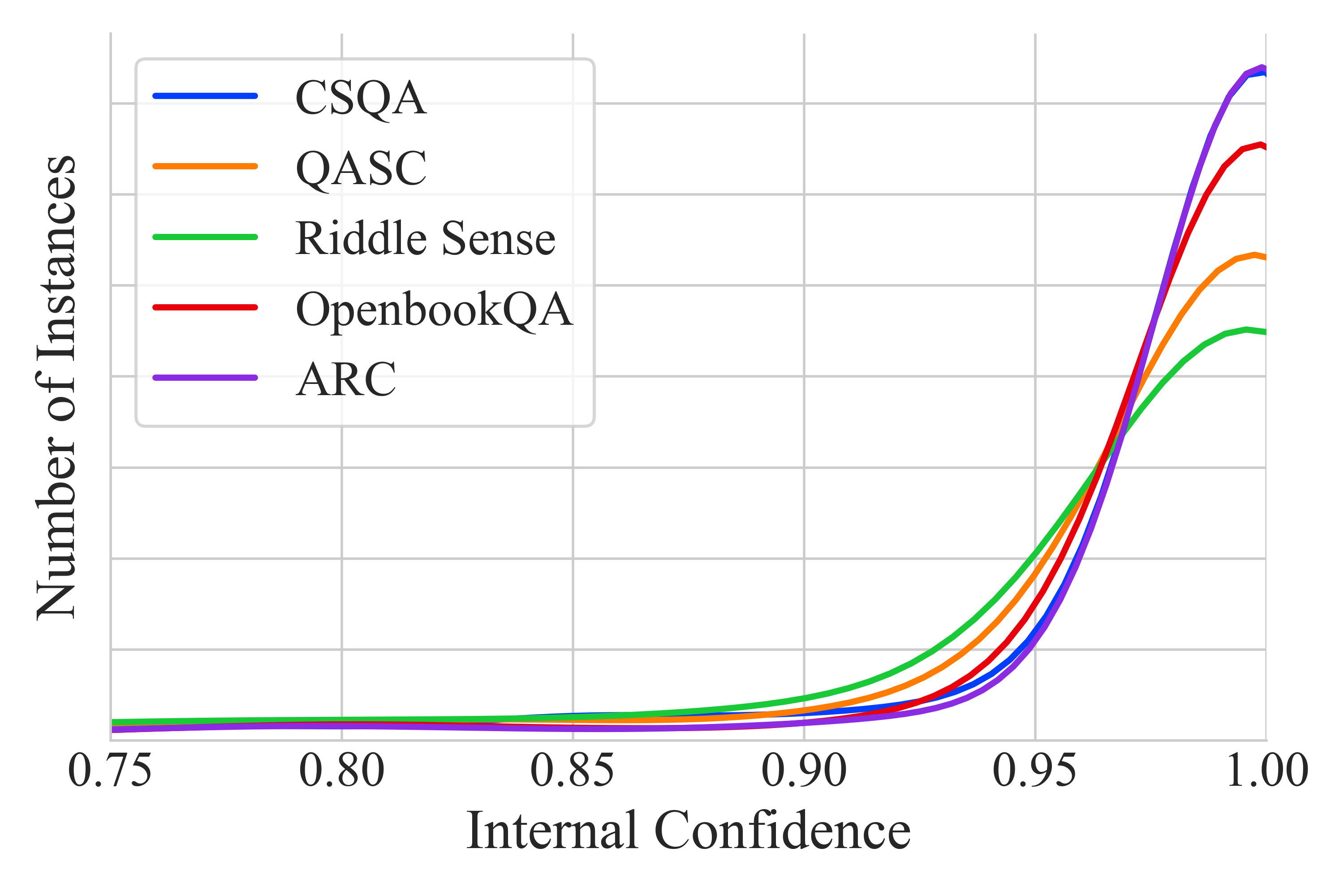}
        
        \caption{The distribution of internal confidence across multiple datasets, composed of the adjusted token probabilities obtained from \textbf{InstructGPT-3 + RLHF}.}
        \label{fig:internal_confidence Distribution GPT3}
    \end{minipage}
\end{figure*}

\setlength{\tabcolsep}{5pt}
\begin{table*}[!htbp]
  \centering
  \newcolumntype{C}{>{\centering\arraybackslash}X}
  \begin{tabularx}{\textwidth}{p{1.6cm} p{1.8cm} | X | c}
    \hline
    \textbf{Int. Conf.} & \textbf{Verb. Cert.} & \textbf{Example} & \textbf{Alignment Type} \\ \hline
     & & \textbf{Q}: What makes someone a nomad? & Consistent Alignment \\
     \textcolor{green}{\up} 1.00 & \textcolor{green}{\up} 1.00 & \textbf{A}: C. have no home & \\
    \hline
     & & \textbf{Q}: Where would using a boat not require navigation skills? & Internal Overconfidence \\
     \textcolor{green}{\up} 0.99 & \textcolor{red}{\down} 0.20 & \textbf{A}: C. garage & \\
    \hline 
     & & \textbf{Q}: What do professors primarily do? & External Overconfidence \\
     \textcolor{red}{\down} 0.66 & \textcolor{green}{\up} 1.00 & \textbf{A}: E. teach courses & \\
    \hline 
     & & \textbf{Q}: Killing people should not cause what emotion? & Consistent Discordance \\
     \textcolor{red}{\down} 0.61 & \textcolor{red}{\down} 0.40 & \textbf{A}: C. joy & \\
    \hline
  \end{tabularx}
  \caption{Instances of similar and contrasting values of internal confidence (Int. Conf.) and verbalized certainty (Verbal Cert.). \textbf{Q} represents the model's question, and \textbf{A} is the model's answer. Tests are from \textbf{InstructGPT-3 + RLHF} (\textit{text-davinci-003}) using the CommonsenseQA dataset.}
  \label{tab:Illustrative Examples GPT3}
\end{table*}

\end{document}